%% file: colm2024_conference.tex
\documentclass{article} 
\usepackage{colm2024_conference}

\usepackage{booktabs}
\usepackage{graphicx}
\usepackage{enumitem}
\usepackage{wrapfig}
\usepackage{algorithm}
\usepackage{algpseudocode}
\usepackage{natbib}
\usepackage{makecell}
\usepackage{booktabs}
\usepackage{bbm}
\usepackage{array}
\usepackage{amsmath}  
\usepackage{amssymb}
\usepackage{amsfonts}
\usepackage{multirow}
\usepackage{verbatim}
\usepackage{caption}
\usepackage{longtable}
\usepackage{supertabular}
\usepackage{hyperref}
\usepackage{CJKutf8}
\usepackage[utf8]{inputenc} 
\usepackage[T1]{fontenc} 
\usepackage[french,vietnamese,mongolian,greek,english]{babel}
\usepackage{pifont}
\usepackage{afterpage}
\usepackage{tablefootnote}
\usepackage{colortbl}
\usepackage{xspace}
\usepackage{textcomp}
\usepackage{makecell}
\usepackage{lscape} 
\usepackage{siunitx}
\usepackage{listings}
\usepackage{xcolor}
\usepackage{adjustbox}
\definecolor{dt}{gray}{0.7}
\definecolor{tongyi-purple}{RGB}{97,92,237}
\colorlet{tongyi-purple-alpha}{tongyi-purple!38}

\usepackage{pifont}       
\usepackage{bbding}       
\usepackage{fontawesome}

\usepackage{scrextend}

\usepackage{tgpagella}
\usepackage{latexsym}
\usepackage[T1]{fontenc}
\usepackage[utf8]{inputenc}
\usepackage{microtype}
\definecolor{mydarkblue}{rgb}{0,0.08,0.45}
\definecolor{citecolor}{HTML}{0071BC}
\usepackage{url}            
\usepackage{nicefrac}       
\usepackage{changepage}
\usepackage{xargs}          
\usepackage{wrapfig,lipsum,booktabs}
\usepackage{longtable}
\usepackage{subcaption}
\usepackage{endnotes}

\usepackage{pgfplots}
\usetikzlibrary{pgfplots.groupplots}
\pgfplotsset{compat=1.3}
\usepackage{tikz}
\usetikzlibrary{patterns}

\usepackage[most]{tcolorbox}
\usepackage{fvextra}
\usepackage{graphicx}
\usepackage[capitalize,noabbrev]{cleveref}
\crefname{section}{Section}{\S\S}
\Crefname{section}{Section}{\S\S}
\crefname{table}{Table}{Tables}
\crefname{figure}{Figure}{Figures}
\crefname{algorithm}{Algorithm}{}
\crefname{equation}{eq.}{}
\crefname{appendix}{Appendix}{}
\crefformat{section}{Section #2#1#3}
\usepackage{multicol}
\usepackage{fancyvrb}
\newsavebox{\myverbcontent}
\usepackage{titlesec}
\titleformat*{\section}{\large\bfseries}

\usepackage{nicematrix} 
\usepackage{arydshln}

\makeatletter
\DeclareRobustCommand\onedot{\futurelet\@let@token\@onedot}
\def\@onedot{\ifx\@let@token.\else.\null\fi\xspace}

\usepackage{todonotes}
\definecolor{tongyiPurple}{HTML}{643CE7}

\usepackage{tabularx}

\RecustomVerbatimEnvironment{verbatim}{Verbatim}{breaklines=true, breakanywhere=true}
\newtcolorbox{promptblock}[1]{
  breakable,
  colback=tongyiPurple!25,      
  colframe=tongyiPurple!75,    
  coltitle=black,       
  fonttitle=\bfseries, 
  title={#1},           
  arc=1mm,              
  boxrule=0.5mm,        
  left=2mm, right=2mm, top=2mm, bottom=2mm, 
  toptitle=1mm, bottomtitle=1mm 
}

\usepackage{soul}

\title{Qwen3-VL-Seg: Unlocking Open-World Referring Segmentation with Vision-Language Grounding}

\author{
Yuan Yao \quad
Qiushi Yang \quad
Humen Zhong \quad
Jiangning Wei \quad
Yifang Men \\
Shuai Bai \quad
Miaomiao Cui \quad
Zhibo Yang \\[2mm]
\textbf{Tongyi Lab, Alibaba Group}
}

\begin{document}

\maketitle


\begin{abstract} 
Open-world referring segmentation requires grounding unconstrained language expressions to precise pixel-level regions. Existing multimodal large language models (MLLMs) exhibit strong open-world visual grounding, but their outputs remain limited to sparse bounding-box coordinates and are insufficient for dense visual prediction. 
Recent MLLM-based segmentation methods either directly predict sparse contour coordinates, struggling to reconstruct continuous object boundaries, or rely on external segmentation foundation models such as the Segment Anything Model (SAM), introducing substantial architectural and deployment overhead.
We present Qwen3-VL-Seg, a parameter-efficient framework that treats the MLLM-predicted box as a semantically grounded structural prior and decodes it into pixel-level referring segmentation. At its core, a lightweight box-guided mask decoder combines multi-scale spatial feature injection, spatial-semantic query construction, box-guided high-resolution pixel fusion, and iterative mask-aware query refinement, introducing only 17M parameters (about 0.4\% of the base model). For scalable open-world training, we construct SA1B-ORS, an SA-1B-derived dataset with two subsets: SA1B-CoRS (category-oriented samples) and SA1B-DeRS (descriptive, instance-specific samples). 
For evaluation, we curate ORS-Bench, a manually screened benchmark with in-distribution and out-of-distribution subsets covering diverse referring expression types. Extensive experiments on referring expression segmentation, visual grounding, and ORS-Bench show that Qwen3-VL-Seg performs strongly across closed-set and open-world settings, with clear advantages on language-intensive instructions and strong out-of-distribution generalization. Evaluations on general multimodal benchmarks further show that the model broadly preserves general-purpose multimodal competence after segmentation-oriented adaptation.
\end{abstract}

\section{Introduction}

The rapid advancement of Multimodal Large Language Models (MLLMs) has significantly expanded the frontier of visual understanding. Beyond image-level question answering and multimodal dialogue, many vision-language tasks require a model to precisely associate linguistic concepts with specific image regions. This capability, commonly referred to as \emph{visual grounding}, forms a critical bridge between linguistic semantics and fine-grained visual evidence.

Recent frontier MLLMs, such as GPT~\citep{hurst2024gpt,singh2025openai}, Gemini~\citep{team2023gemini,comanici2025gemini,gemini3_flash_model_card,gemini3_pro_tech_report}, and Qwen-VL~\citep{Qwen2-VL,Qwen2.5-VL,Qwen3-VL}, have demonstrated substantial progress in visual grounding. Through large-scale instruction tuning, these models can localize arbitrary visual entities with natural language instructions, enabling them to move beyond closed-set vocabularies. However, their outputs are still typically limited to bounding boxes, which provide only coarse spatial localization. Such box-level grounding is insufficient for applications requiring precise boundary-aware spatial reasoning, including robotic manipulation, medical image analysis, and fine-grained image editing. A natural next step is therefore to move from box-level localization to pixel-level mask prediction, namely \emph{referring segmentation}.

Existing approaches to extending MLLMs from box-level grounding to pixel-level referring segmentation mainly follow two technical routes. The first couple of MLLMs with external segmentation models, most commonly SAM~\citep{kirillov2023segany}. For example, LISA~\citep{lai2024lisa}, GSVA~\citep{xia2024gsva}, and SAM4MLLM~\citep{chen2024sam4mllm} use MLLM representations to identify target regions and then invoke SAM or SAM-like decoders to generate masks. Although effective, these decoupled pipelines introduce nontrivial parameter overhead, additional architectural dependencies, and less streamlined deployment. Separately, SAM3~\citep{carion2025sam} studies concept-prompted segmentation within a segmentation foundation model rather than an MLLM-plus-decoder formulation. The second route avoids external segmenters by equipping MLLMs with lightweight dense prediction heads or autoregressive mask representations. Examples include Text4Seg~\citep{lan2024text4seg}, UFO~\citep{tang2025ufo}, and recent lightweight mask-decoder approaches such as MLLMSeg~\citep{wang2025unlocking}. However, text-based formulations often suffer from discretization artifacts or representation bottlenecks, while lightweight dense decoders still struggle to recover precise boundaries from low-resolution latent features alone.

Open-world referring segmentation (ORS) remains under-explored in existing benchmarks and methods. In terms of benchmarks, dominant datasets remain constrained by closed-set vocabularies. Classic Referring Expression Segmentation (RES)~\citep{hu2016segmentation} benchmarks are built around a relatively limited object space, while Generalized RES (GRES)~\citep{liu2023gres} broadens the scope to include multi-target and no-target cases, and ReasonSeg~\citep{lai2024lisa} advances the paradigm by incorporating reasoning-intensive instructions. Despite these progresses, existing benchmarks fail to fully leverage the open-vocabulary capabilities inherent in modern MLLMs to achieve open-world training and evaluation. Methodologically, existing approaches either rely on heavyweight external segmentation components or still struggle to recover precise boundaries when using compact dense predictors. Moreover, most methods do not directly exploit the grounding priors already present in modern MLLMs. This raises a central question: \emph{can we transform the open-world grounding capability of MLLMs into pixel-level referring segmentation without relying on heavyweight external segmentation models?}

To address these limitations, we introduce \textbf{Qwen3-VL-Seg}, a parameter-efficient framework that converts box-level grounding priors of a pretrained MLLM into pixel-level referring segmentation without relying on an external segmentation model. Our key insight is that the bounding box predicted by the MLLM is not merely a terminal output, but a semantically grounded and spatially informative anchor for mask decoding. Consequently, we leverage the box as a structural prior throughout the decoder. Specifically, we design a lightweight box-guided mask decoder with four components: (i) \emph{multi-scale spatial feature injection}, which enriches intermediate visual features for dense prediction; (ii) \emph{spatial-semantic query construction}, which fuses box geometry and language features into an object query; (iii) \emph{box-guided high-resolution pixel fusion}, which converts the predicted box into a differentiable soft gate to inject fine-grained image textures while suppressing background clutter; and (iv) \emph{iterative mask-aware query refinement}, which feeds first-pass mask evidence back into the query to progressively sharpen object boundaries. In this way, box-level grounding is systematically transformed into mask-level prediction while introducing only 17M additional parameters, accounting for approximately 0.4\% of the base MLLM.

Towards model training, we utilize existing public datasets including the RefCOCO series~\citep{kazemzadeh2014referitgame,yu2016modeling,mao2016generation}, LVIS~\citep{gupta2019lvis}, and COCO~\citep{lin2014microsoft} as supervised sources. To further scale open-world referring segmentation, we construct \textbf{SA1B-ORS} (\textbf{O}pen-world \textbf{R}eferring \textbf{S}egmentation), an SA-1B-derived dataset built from a sampled pool of 2 million raw SA-1B images. SA1B-ORS is a composite dataset with two complementary subsets: \textbf{SA1B-CoRS} (Category-oriented Referring Segmentation), which converts category-agnostic mask fragments into category-level referring supervision where one expression may refer to one or multiple instances of the same semantic class; and \textbf{SA1B-DeRS} (Descriptive Referring Segmentation), which provides instance-specific supervision with attributes, relations, or contextual cues so that a single target can be uniquely identified. In total, SA1B-ORS contains 1.05 million SA1B-CoRS samples and 1.94 million SA1B-DeRS samples, providing both broad category-level coverage and fine-grained descriptive supervision.

To systematically evaluate model generalization, we introduce \textbf{ORS-Bench}, a benchmark suite containing an in-distribution subset (\textbf{ORS-ID-Bench}) and an out-of-distribution subset (\textbf{ORS-OOD-Bench}). ORS-ID-Bench includes 9,055 manually curated samples across four instruction formats, providing a reliable assessment aligned with the training distribution. ORS-OOD-Bench is designed to probe the limits of model capability and robustness under challenging scenarios by covering six distributional shifts: category, instance scale, instruction complexity, occlusion, lighting, and domain risk. Guided by training distribution analysis, we curate approximately 200 challenging samples per dimension from non-overlapping sources, spanning extreme spatial extents, indirect reasoning, adverse environments, and safety-critical domains such as autonomous driving and medical diagnosis. ORS-Bench thus offers a rigorous testbed for measuring the operational limits of referring segmentation models.

Comprehensive experiments on referring expression segmentation and visual grounding on both public and our curated benchmarks show that Qwen3-VL-Seg performs strongly across both closed-set and open-world settings. The gains are especially pronounced on language-intensive open-world instructions, where precise alignment between semantic understanding and mask prediction is most critical. In addition, the model demonstrates strong out-of-distribution generalization on ORS-OOD-Bench, while additional general multimodal evaluation shows that it broadly preserves general-purpose multimodal competence after segmentation-oriented adaptation.

In summary, our main contributions are as follows:

\begin{enumerate}[leftmargin=*]
    \item We introduce Qwen3-VL-Seg, a parameter-efficient framework that converts the open-world grounding priors of pretrained MLLMs into pixel-level referring segmentation without relying on external foundation models such as SAM.
    
    \item We propose a lightweight box-guided mask decoder that leverages MLLM-predicted boxes as structural priors and transforms them into precise masks through multi-scale spatial feature injection, spatial-semantic query construction, box-guided high-resolution pixel fusion, and iterative mask-aware query refinement.
    
    \item We construct SA1B-ORS, an SA-1B-derived open-world referring segmentation dataset with two complementary subsets: SA1B-CoRS for category-oriented references and SA1B-DeRS for fine-grained descriptive references.
    
    \item We introduce ORS-Bench, comprising ORS-ID-Bench and ORS-OOD-Bench, to evaluate open-world referring segmentation under both in-distribution formats and six distinct out-of-distribution shifts.
    
    \item We conduct extensive experiments on referring expression segmentation, grounding, open-world referring segmentation, and general multimodal evaluation benchmarks, showing that Qwen3-VL-Seg achieves strong performance across both closed-set and open-world settings, with especially clear gains on language-intensive instructions, strong out-of-distribution generalization, and broad preservation of general-purpose multimodal competence.
\end{enumerate}

\section{Related Work}

\paragraph{Referring Expression Segmentation.}
Referring Expression Segmentation (RES) studies pixel-level localization of objects specified by language expressions, with classic benchmarks largely built around the one-expression, one-instance setting. Generalized RES (GRES) extends this formulation to multi-target and no-target cases through the gRefCOCO benchmark, while LISA~\citep{lai2024lisa} broadens the instruction space toward reasoning-intensive segmentation. These works enrich the expression space of language-guided segmentation, but the dominant benchmarks remain largely human-curated and closed-vocabulary, limiting scalability to open-world entity spaces.

\paragraph{MLLMs for region understanding and segmentation.}
A parallel line of work equips MLLMs with explicit region grounding ability. Representative models such as Shikra~\citep{chen2023shikra} and Ferret~\citep{you2023ferret} support open-vocabulary region understanding, while more recent systems such as Youtu-VL~\citep{wei2026youtu} suggest that vision-centric supervision can be absorbed into standard MLLM training at scale. Building on these advances, recent methods extend MLLMs from region grounding to pixel-level prediction. SAM-based approaches, such as LISA, GSVA, and SAM4MLLM~\citep{chen2024sam4mllm}, couple MLLMs with external segmentation models to obtain high-quality masks, but inherit the corresponding parameter and deployment cost. Recent progress on promptable concept segmentation, exemplified by SAM3~\citep{carion2025sam}, further strengthens this external-model line by extending large segmentation foundation models to concept-prompted segmentation, detection, and tracking. These models provide strong open-world mask prediction capability, but remain architecturally distinct from MLLM-native referring segmentation and preserve the deployment cost of large external segmenters. SAM-free approaches instead rely on lightweight dense heads or unified autoregressive formulations, including PerceptionGPT~\citep{pi2024perceptiongpt}, Text4Seg~\citep{lan2024text4seg}, UFO~\citep{tang2025ufo}, and MLLMSeg~\citep{wang2025unlocking}. Our approach is also lightweight and SAM-free at inference, but uses a different inductive bias: the MLLM-predicted box is treated as a structural prior for query construction, pixel fusion, and iterative refinement.

\paragraph{Data for open-world referring segmentation.}
Existing referring segmentation datasets, including RefCOCO series~\citep{kazemzadeh2014referitgame,yu2016modeling,mao2016generation}, gRefCOCO~\citep{liu2023gres}, and ReasonSeg~\citep{lai2024lisa}, have been crucial to progress, but they are limited in vocabulary, scale, or construction cost. Our work focuses on scalable construction of open-world referring segmentation data from SA-1B~\citep{kirillov2023segany}. SA1B-ORS is organized into two complementary subsets: SA1B-CoRS converts fragmented category-agnostic masks into category-oriented referring samples through entity distillation, mask reconstruction, MLLM verification, and caption generation, while SA1B-DeRS provides descriptive instance-level instructions with grounding-based filtering. This dataset organization aligns training supervision with the open-world grounding regime of modern MLLMs.

\begin{figure}[t]
\centering
\includegraphics[width= 0.9\linewidth]{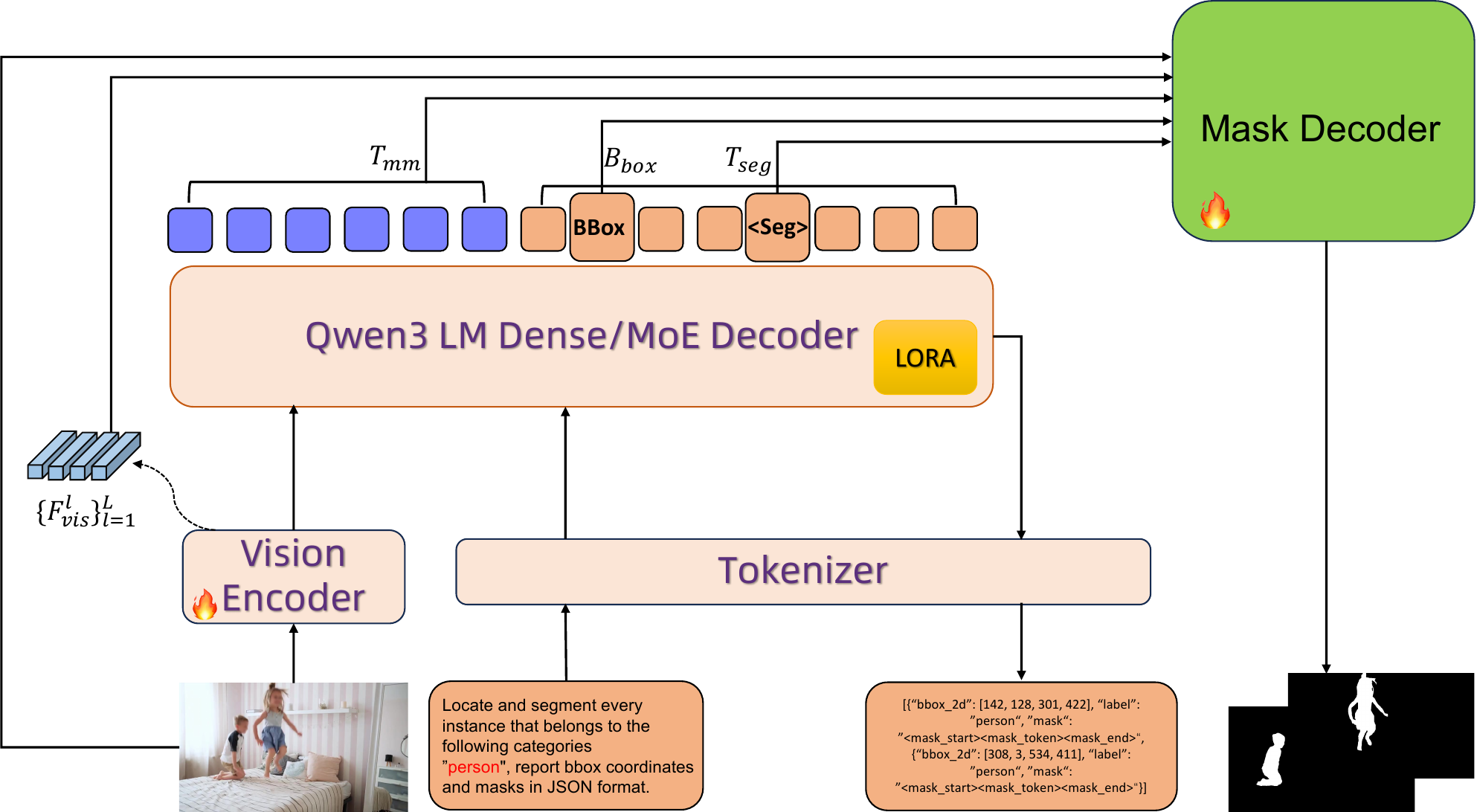}
   \caption{
   Overview of the Qwen3-VL-Seg architecture. 
}
\label{fig:arc}
\end{figure}

\section{Method}

\subsection{Overview}

Open-world referring segmentation aims to ground natural language expressions to specific image regions and generate corresponding pixel-level masks. Modern multimodal large language models (MLLMs) exhibit strong open-world grounding ability, but their outputs are typically limited to coarse localization, e.g., bounding boxes. This creates a mismatch between grounding and segmentation: grounding provides coarse spatial localization, whereas segmentation requires precise pixel-level delineation.

To bridge this gap, we propose a box-guided mask decoder architected upon a pretrained MLLM, as illustrated in Figure~\ref{fig:arc}. Our key insight is to leverage the grounded bounding box as a structural prior for dense mask decoding, rather than treating it as a mere auxiliary prediction. Concretely, the decoder first enriches intermediate ViT features with lightweight spatial adapters and combines them with multimodal visual embeddings to form a dense memory representation. It then constructs object queries by jointly encoding language semantics and box geometry. To recover fine boundaries, we further introduce a box-guided high-resolution pixel fusion module, which softly gates shallow image features before fusing them with upsampled visual features. Finally, we perform iterative mask-aware query refinement by pooling target-aware pixel evidence from the first-pass mask and feeding it back to the query for a second-pass prediction.

Overall, the proposed decoder follows a coarse-to-fine design: the pretrained MLLM provides robust grounding priors, and the box-guided decoder progressively transforms these priors into fine-grained segmentation masks.

\begin{figure}[t]
\centering
\includegraphics[width= 0.9\linewidth]{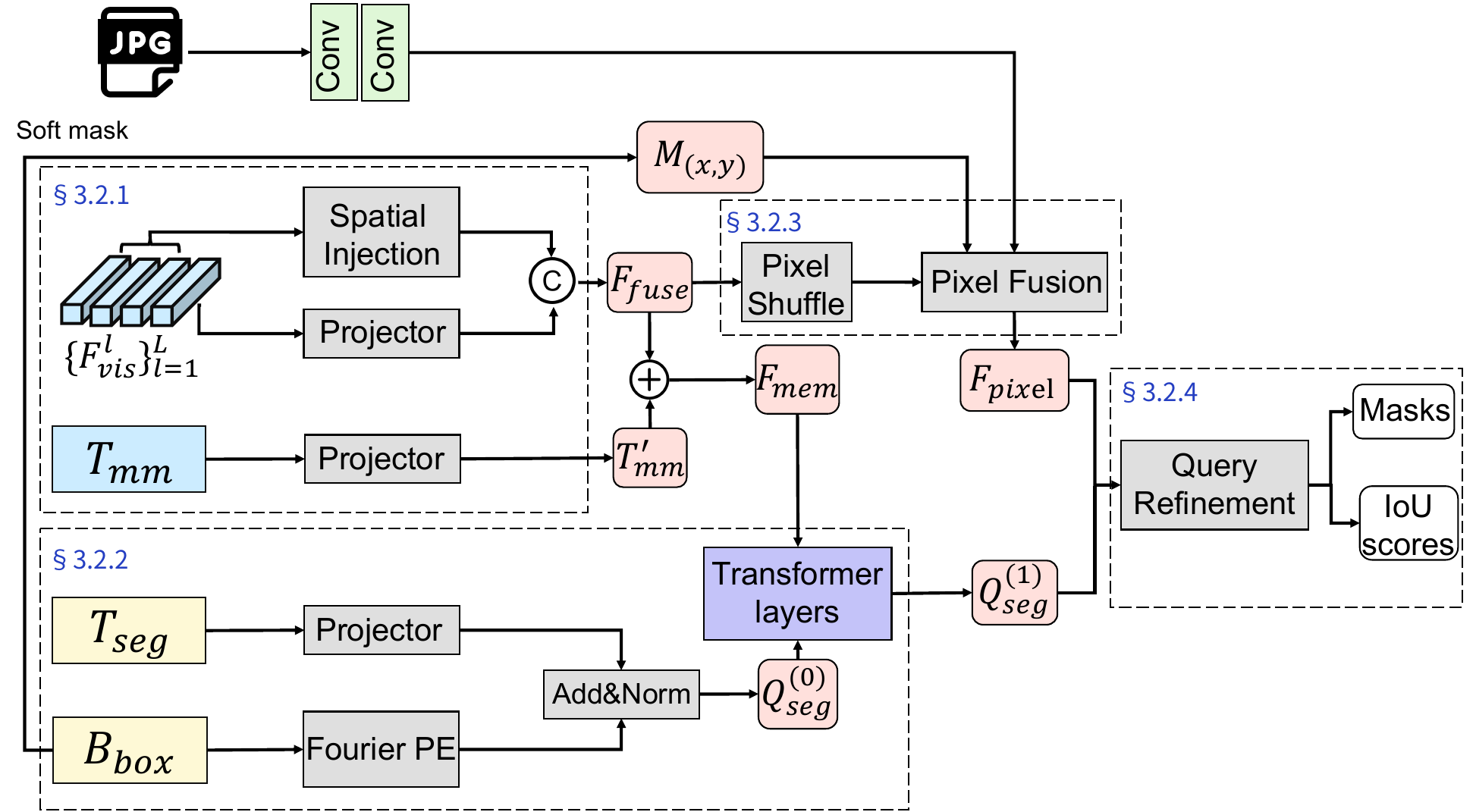}
   \caption{
   Detailed structure of the lightweight mask decoder.
}
\label{fig:decoder}
\end{figure}

Given an input image $I$ and a referring expression, the pretrained MLLM provides four types of information. These include multi-scale visual features $\{F_{\mathrm{vis}}^{l}\}_{l=1}^{L}$, where the $L$-th feature denotes the top-layer visual representation; multimodal visual embeddings $T_{\mathrm{mm}}$; text-conditioned segmentation token features $T_{\mathrm{seg}}$; and a grounded bounding box $B_{\mathrm{box}}$. Our decoder predicts the final mask:
\begin{equation}
\hat{M}=\mathcal{D}\!\left(\{F_{\mathrm{vis}}^{l}\}_{l=1}^{L}, T_{\mathrm{mm}}, T_{\mathrm{seg}}, B_{\mathrm{box}}, I\right)
\end{equation}
where $\mathcal{D}$ denotes the proposed box-guided decoder. As illustrated in Figure~\ref{fig:decoder}, the decoder consists of four components: multi-scale spatial feature injection, spatial-semantic query construction, box-guided high-resolution pixel fusion, and iterative mask-aware query refinement.

\subsection{Model Architecture}

\subsubsection{Multi-scale Spatial Feature Injection and Memory Construction}

High-level MLLM features are typically too coarse to support accurate boundary delineation. We therefore extract intermediate visual features and adapt them with a lightweight \texttt{SpatialFeatureInjector}. For each feature map $F_{\mathrm{vis}}^{l}$, we first project it into the decoder hidden dimension:
\begin{equation}
X_{0}^{(l)}=\mathrm{Conv}_{1\times1}(F_{\mathrm{vis}}^{l})
\end{equation}
We then inject local spatial bias through a depthwise branch:
\begin{equation}
\tilde{F}_l = X_{0}^{(l)} + s \cdot \mathrm{GELU}\!\left(\mathrm{DWConv}\!\left(\mathrm{GroupNorm}(X_{0}^{(l)})\right)\right)
\end{equation}
where $s$ is a learnable scalar initialized to $10^{-3}$. This near-zero initialization keeps the adapter close to an identity mapping at the start of fine-tuning, which stabilizes optimization. The adapted features $\{\tilde{F}_{l}\}_{l=1}^{L-1}$ and the projected top-layer feature $\tilde{F}_{L}$ are concatenated and fused by a lightweight convolutional fusion block:
\begin{equation}
F_{\mathrm{fuse}}=\phi_{\mathrm{fuse}}\!\left(\tilde{F}_1 \oplus \cdots \oplus \tilde{F}_{L-1} \oplus \tilde{F}_{L}\right)
\end{equation}

To construct the decoder memory, we project the multimodal visual embeddings and reshape them into a 2D feature map:
\begin{equation}
T'_{\mathrm{mm}}=\mathrm{Reshape}\!\left(W_{\mathrm{mm}}T_{\mathrm{mm}}\right).
\end{equation}
The final memory feature is obtained as:
\begin{equation}
F_{\mathrm{mem}} = T'_{\mathrm{mm}} + F_{\mathrm{fuse}} + P_{\mathrm{mem}}
\end{equation}
where $P_{\mathrm{mem}}$ denotes a learnable 2D positional encoding and is omitted in Figure~\ref{fig:decoder} for clarity. Flattening $F_{\mathrm{mem}}$ yields the transformer decoder memory, which combines language-aligned multimodal semantics with spatially enhanced visual features, providing a dense representation for mask prediction.

\subsubsection{Spatial-Semantic Query Construction}

The object query should capture both target semantics and instance-level spatial identity. We therefore use the grounded box as an explicit conditioning signal during query construction.

Let the grounded box be:
\begin{equation}
B_{\mathrm{box}}=(x_{1},y_{1},x_{2},y_{2})
\end{equation}
We obtain the width $w$, height $h$ from the box coordinates and encode its geometry with Fourier positional encoding $\gamma(\cdot)$:
\begin{equation}
E_{\mathrm{box}}=\gamma(x_{1})\oplus\gamma(y_{1})\oplus\gamma(0.2\log w+0.5)\oplus\gamma(0.2\log h+0.5)
\end{equation}
where the log-scale transform improves robustness to large variations in object scale. The initial object query is then defined as:
\begin{equation}
Q_{\mathrm{seg}}^{(0)}=\mathrm{LayerNorm}\!\left(\mathrm{MLP}_{\mathrm{box}}(E_{\mathrm{box}})+W_{\mathrm{seg}}T_{\mathrm{seg}}\right)
\end{equation}
A stack of transformer decoder layers attends to the global memory and produces decoded query features:
\begin{equation}
Q_{\mathrm{seg}}^{(1)}=\mathrm{Decoder}(Q_{\mathrm{seg}}^{(0)},F_{\mathrm{mem}})
\end{equation}
As a result, the query is jointly conditioned on target semantics and instance-level spatial cues. Rather than relying on language alone to retrieve the target, the decoder starts from a query that already carries a grounded spatial prior.

\subsubsection{Box-Guided High-Resolution Pixel Fusion}

To recover boundary details that are largely absent from high-level MLLM features, we extract shallow image features through a lightweight convolutional stem:
\begin{equation}
F_{\mathrm{cnn}}=\mathrm{Stem}(I)
\end{equation}
Directly fusing such shallow features may introduce substantial background clutter. We therefore use the grounded box to construct a soft spatial gate.

Given the box coordinates $(x_1,y_1,x_2,y_2)$, we enlarge the box by $15\%$ in width and height to tolerate localization errors, producing $(x'_1,y'_1,x'_2,y'_2)$. The resulting gate is
\begin{equation}
M(x,y)=\sigma(\alpha(x-x'_1))\cdot\sigma(\alpha(x'_2-x))\cdot\sigma(\alpha(y-y'_1))\cdot\sigma(\alpha(y'_2-y))
\end{equation}
where $\sigma(\cdot)$ is the sigmoid function and $\alpha=20$. When multiple boxes are available, we take the spatial maximum over their gates.

Meanwhile, the fused visual feature is progressively upsampled by a two-stage PixelShuffle module:
\begin{equation}
F_{\mathrm{up}}=\mathrm{Upsample}(F_{\mathrm{fuse}})
\end{equation}
We then fuse the gated shallow features with the upsampled visual feature:
\begin{equation}
F_{\mathrm{pixel}}=F_{\mathrm{up}} \oplus (M(x,y)\odot F_{\mathrm{cnn}})
\end{equation}
where $\odot$ denotes element-wise multiplication. This design injects high-frequency local details into the decoder while suppressing irrelevant responses outside the grounded region, producing a detail-rich pixel representation with reduced background interference.

\subsubsection{Iterative Mask-Aware Query Refinement}

A single-round mask prediction is often insufficient for thin structures or cluttered scenes. We therefore introduce a lightweight refinement step that feeds mask-aware pixel evidence back into the query.

In the first pass, the decoded query $Q_{\mathrm{seg}}^{(1)}$ generates dynamic kernels to predict an initial mask:
\begin{equation}
M_{\mathrm{logit}}^{(1)}=\Psi(Q_{\mathrm{seg}}^{(1)},F_{\mathrm{pixel}})
\end{equation}
where $\Psi(\cdot)$ denotes dynamic mask prediction over the pixel feature map. We then use the corresponding soft mask to pool a target-aware feature:
\begin{equation}
F_{\mathrm{tar}}=
\frac{\sum_{h,w}\left(\sigma(M_{\mathrm{logit}}^{(1)})\odot F_{\mathrm{pixel}}\right)}
{\sum_{h,w}\sigma(M_{\mathrm{logit}}^{(1)})+\epsilon}
\end{equation}
where $\epsilon=10^{-6}$ ensures numerical stability.

The pooled feature is projected and added back to the decoded query:
\begin{equation}
Q_{\mathrm{seg}}^{(2)}=\mathrm{LayerNorm}\!\left(Q_{\mathrm{seg}}^{(1)}+\phi_{\mathrm{ref}}(F_{\mathrm{tar}})\right)
\end{equation}
The refined query is then used for second-pass mask prediction:
\begin{equation}
M_{\mathrm{logit}}^{(2)}=\Psi(Q_{\mathrm{seg}}^{(2)},F_{\mathrm{pixel}})
\end{equation}
Finally, the mask logits are upsampled to the target resolution:
\begin{equation}
\hat{M}=\mathrm{Interp}(M_{\mathrm{logit}}^{(2)})
\end{equation}
In parallel, an auxiliary IoU head predicts the mask confidence from the refined query:
\begin{equation}
\hat{s}_{\mathrm{IoU}}=W_{\mathrm{IoU}}Q_{\mathrm{seg}}^{(2)}
\end{equation}
This refinement loop establishes an explicit interaction between query prediction and pixel evidence, allowing the decoder to correct coarse initial masks and sharpen object boundaries.

\section{Dataset and Benchmark Construction}

\subsection{SA1B-ORS}

\subsubsection{Overview}

\begin{figure}[t]
\centering
\includegraphics[width=1\linewidth]{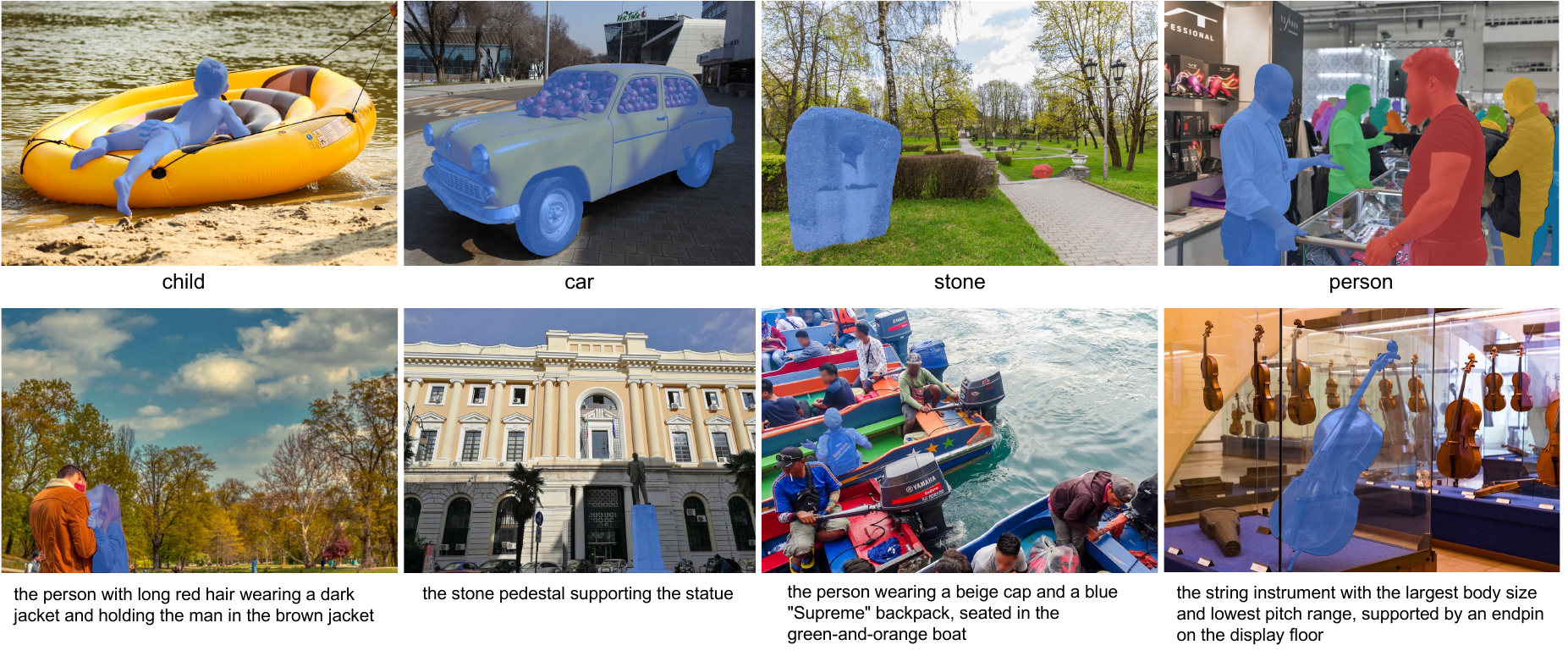}
\caption{Examples from SA1B-ORS. The first row shows SA1B-CoRS, where each category-oriented expression corresponds to one or multiple entities. The second row shows SA1B-DeRS, where each descriptive instruction corresponds to a single entity.}
\label{fig:dataset}
\end{figure}

To support open-world referring segmentation, we construct \textbf{SA1B-ORS}, a large-scale composite dataset derived from 2 million raw images sampled from SA-1B. SA1B-ORS contains two complementary subsets: \textbf{SA1B-CoRS} (Category-oriented Referring Segmentation) and \textbf{SA1B-DeRS} (Descriptive Referring Segmentation), yielding 1.05M and 1.94M samples, respectively. SA1B-CoRS provides scalable category-consistent supervision, where each expression refers to a single semantic category and the target may contain one or multiple entities of that category. SA1B-DeRS complements this setting with descriptive, instance-specific supervision for scenes in which a category name alone is insufficient to disambiguate the target. Together, the two subsets provide both broad category-level coverage and fine-grained instance-level supervision. The selected examples are shown in Figure~\ref{fig:dataset}.

\subsubsection{SA1B-CoRS}

SA-1B provides large-scale masks, but its annotations are fragmented, category-agnostic, and not paired with language. SA1B-CoRS converts this raw supervision into category-oriented referring segmentation data through five stages: \emph{instance distillation}, \emph{coarse mask acquisition}, \emph{fine mask merging}, \emph{MLLM verification}, and \emph{referring caption generation}. In SA1B-CoRS, each expression refers to a single semantic category, while the target may contain one or multiple entities of that category.

\paragraph{Instance Distillation.}

We first identify reliable referable entities in each image. Since many SA-1B fragments correspond to partial regions, amorphous structures, or non-instance concepts, we begin with open-vocabulary tags from RAM++~\citep{zhang2023recognize} and apply hybrid semantic filtering. A label is retained if it belongs to a curated high-priority vocabulary or can be mapped, via WordNet hypernym paths, to referable semantic roots such as \textit{person}, \textit{vehicle}, \textit{animal}, \textit{artifact}, \textit{clothing}, \textit{food}, and \textit{structure}. Scene words, colors, events, body parts, and other weakly referable concepts are removed.

The remaining labels are then verified by an open-vocabulary grounding model. We apply hierarchical non-maximum suppression to preserve specific labels among semantically overlapping concepts while allowing different categories with spatial overlap to coexist. This yields a visually validated entity set:
\begin{equation}
\mathcal{O}=\{(l_i, b_i)\}_{i=1}^{N}
\end{equation}
where $l_i$ and $b_i$ denote the entity label and grounded box, respectively.

\paragraph{Coarse Mask Acquisition.}

Given the distilled entity set, we construct an initial mask for each target entity or entity set. For each label-image pair, Qwen3-VL-Plus~\citep{Qwen3-VL} predicts a grounded box and SAM2~\citep{ravi2024sam2} produces a coarse mask, thereby converting box-level grounding into a mask-level entity hypothesis. This coarse mask is used only as an entity prior to subsequent refinement.

\paragraph{Fine Mask Merging.}

SA-1B often splits a single semantic entity or entity set into multiple partial masks. To recover an entity-level mask, we merge SA-1B fragments under the guidance of the coarse mask.

Let $m_p$ denote the coarse mask and $\{m_f^k\}$ the SA-1B fragment masks from the same image. We first discard very small fragments and fragments whose bounding boxes do not intersect the coarse-mask region. For each remaining fragment, we compute
\begin{equation}
\mathrm{IoF}(m_p, m_f^k)=\frac{|m_p \cap m_f^k|}{|m_f^k|}
\end{equation}
A fragment is selected if its overlap with the coarse mask is sufficiently high. We use IoF rather than IoU because the key question is whether a small SA-1B fragment belongs to the target entity.

Starting from the selected fragments, we merge spatially supported fragments into an entity-level mask and recover nearby uncovered regions from the coarse mask to compensate for incomplete SA-1B coverage. The merged result is further refined by hole filling, morphological cleaning, and connected-component filtering. If refinement causes a substantial area drop, we fall back to the coarse mask. We also remove duplicate candidates for the same label using box and mask overlap. This coarse-to-fine process transforms fragmented SA-1B annotations into cleaner and more complete entity-level masks.

\begin{figure}[t]
\centering
\includegraphics[width=1\linewidth]{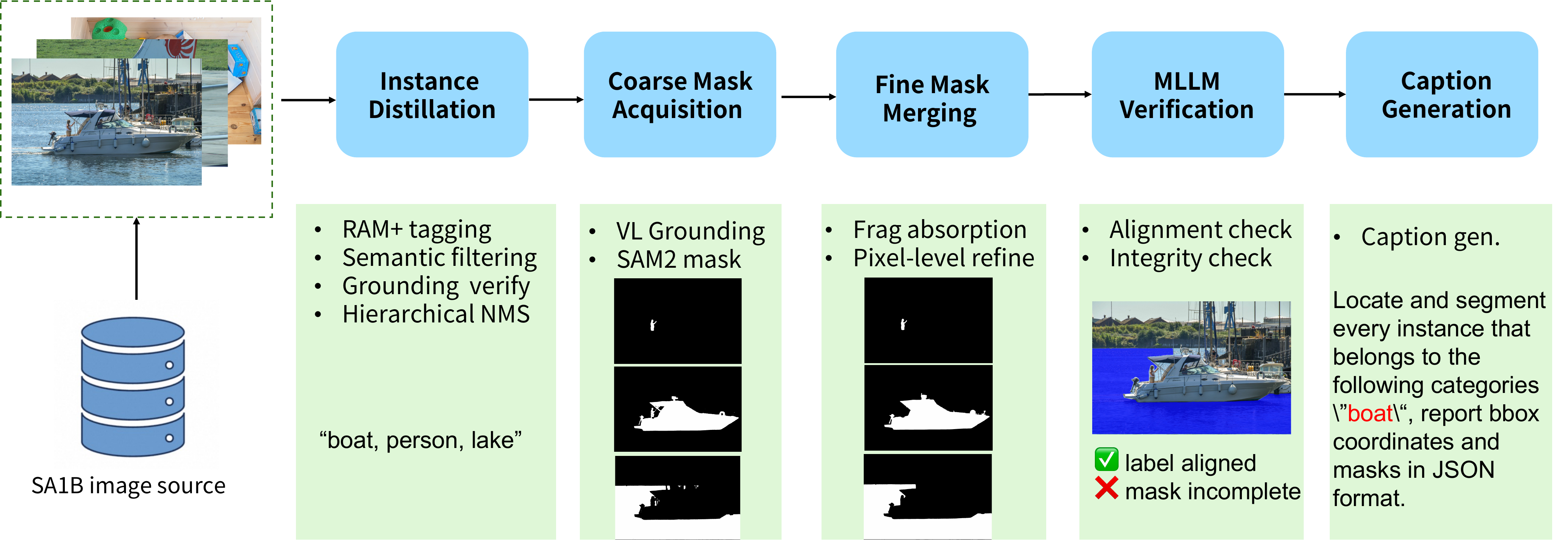}
\caption{Illustration of the curation pipeline for SA1B-CoRS.}
\label{fig:cors}
\end{figure}

\paragraph{MLLM Verification.}

Although mask merging substantially improves the raw SA-1B fragments, the resulting masks may still contain category mismatch, insufficient coverage, contamination from nearby regions, fragmentation, or omission of salient target instances. We therefore introduce an MLLM-based verification stage before caption generation.

For each candidate entity, we construct a verification triplet consisting of the original image, a mask-rendered image in which the candidate region is highlighted in blue, and the corresponding candidate label. This triplet is fed into Qwen3-VL-Plus, which assesses whether the mask is sufficiently reliable to serve as supervision for the given label. The verifier checks category consistency, target coverage, non-target contamination, fragmentation, and omission of salient target regions. The criterion is designed for training usability rather than perfect pixel-level completeness: minor boundary inaccuracies are tolerated, whereas masks with substantial omissions, recognizable non-target regions, severe fragmentation, or missing similarly prominent nearby instances are rejected. Only verified candidates are retained for caption generation.

\paragraph{Referring Caption Generation.}

After MLLM verification, we convert each retained entity annotation into a language-grounded training sample. For each verified annotation, we use its validated mask, associated label, and grounded box to generate a referring expression paired with the image and pixel-level annotation. Each SA1B-CoRS sample contains an image, a referring expression, the corresponding target mask, and metadata such as category and bounding box. After the full pipeline, SA1B-CoRS contains 1.05M training samples.

SA1B-CoRS provides scalable category-oriented supervision, but it does not fully cover cases in which multiple instances of the same category must be distinguished by attributes, relations, or context. This motivates SA1B-DeRS, a complementary subset for descriptive referring segmentation.

\subsubsection{SA1B-DeRS}

\begin{figure}[t]
\centering
\includegraphics[width=1\linewidth]{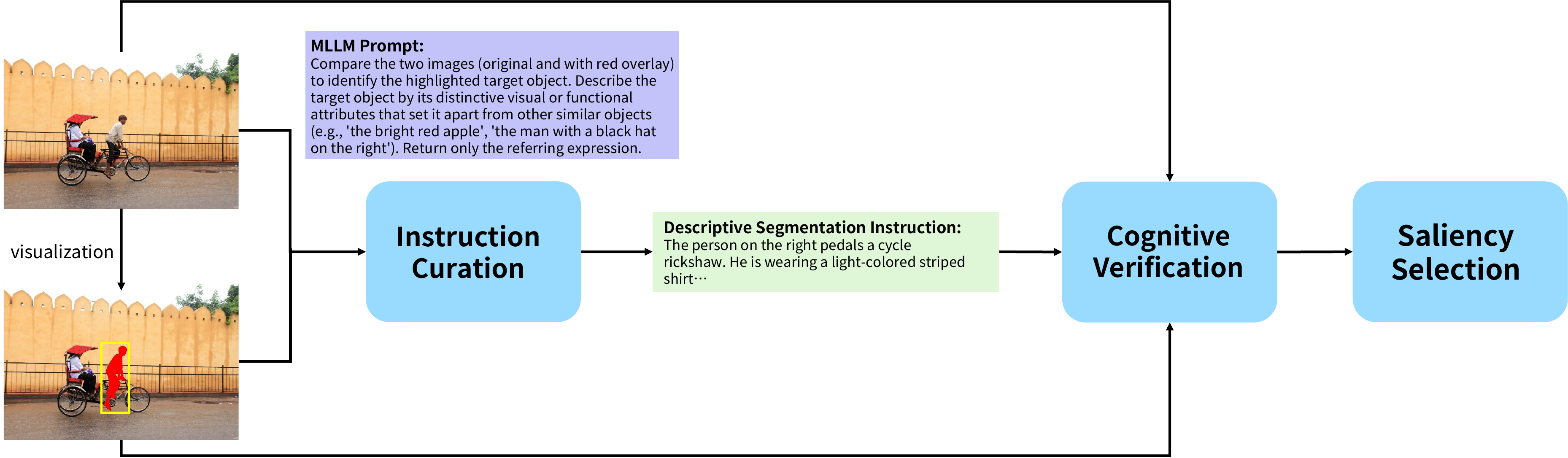}
\caption{Illustration of the curation pipeline for SA1B-DeRS.}
\label{fig:dees}
\end{figure}

To address complex scenarios where category names alone are insufficient for unique instance identification, we introduce SA1B-DeRS, which focuses on descriptive referring segmentation. In scenes with multiple visually similar instances, unambiguous reference requires descriptive cues such as spatial relationships or specific attributes. Since publicly available datasets for this task are scarce, as illustrated in Figure~\ref{fig:dees}, we develop an automated pipeline based on the SA1B dataset to generate high-quality descriptive instructions. The construction process consists of three sequential stages: instruction curation, cognitive verification and saliency selection.

\paragraph{Instruction Curation.}

We begin by leveraging raw images from the SA1B dataset. For each image, we identify instances of interest and extract their ground-truth bounding box coordinates. To generate rich and discriminative descriptions, we feed the original image, a version with the target mask overlaid, and the instance coordinates into the Qwen3-VL-Plus model. We prompt the MLLM to perform a comparative analysis of the visual inputs and generate descriptive instructions from five key perspectives: category name, target attributes, target state, relative position, and contextual relations. This multi-faceted approach ensures that the generated instructions highlight distinctive characteristics, effectively distinguishing the target from other similar objects in cluttered scenes.

\paragraph{Cognitive Verification.}
To ensure the reliability and precision of the generated instructions, we implement a rigorous cognitive filtering mechanism based on MLLM grounding capabilities. The original image and the generated descriptive instruction are fed back into the Qwen3-VL-Plus model, which is tasked with localizing the target referred to by the text and outputting predicted bounding box coordinates. We then compute the IoU between these MLLM-predicted coordinates and the ground-truth bounding boxes. Samples with an IoU below the threshold (0.8 in this work) are discarded, as this threshold indicates a failure in accurate grounding. This step serves as a critical quality control measure, ensuring that the descriptive instruction is not only semantically coherent but also precisely aligned with the visual target, thereby filtering out ambiguous or inaccurate referring expressions.

\paragraph{Saliency Selection.}
Finally, we apply a saliency-based filter to ensure the selected instances are visually prominent and semantically meaningful. We analyze the distribution of the area ratio between each instance’s segmentation mask and the full image across the dataset. Instances with excessively small area ratios, which often correspond to fragmented, trivial, or noisy segments, are discarded. By retaining only those instances with significant spatial extents, we ensure that the dataset focuses on salient objects that are critical for robust visual perception, reducing noise introduced by negligible background elements. The resulting SA1B-DeRS dataset comprises salient targets paired with accurate, highly discriminative descriptive instructions, serving as a crucial component for enhancing models' descriptive referring segmentation capabilities.

\begin{figure}[t]
\centering
\includegraphics[width=1\linewidth]{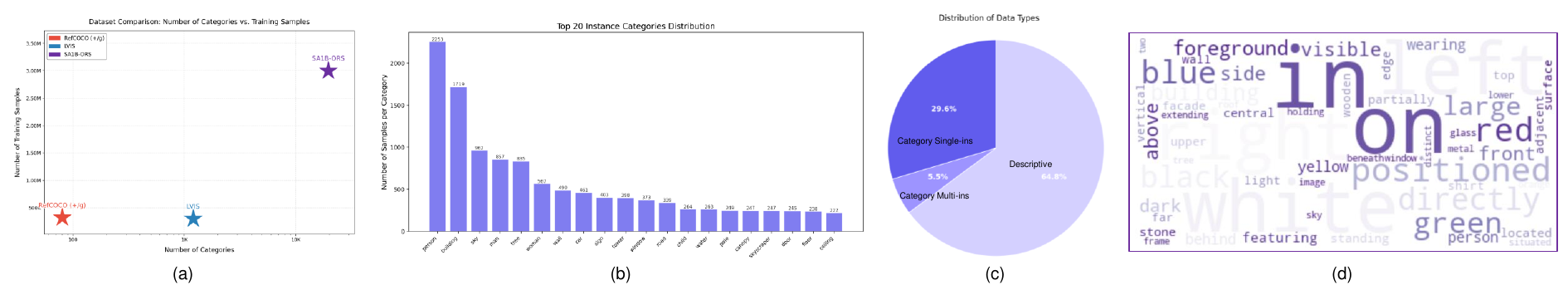}
\caption{Statistical analysis of the SA1B-ORS. (a) Comparison of data scale and category diversity against RefCOCO and LVIS, highlighting the superior size and richness of SA1B-ORS. (b) Distribution of the top-20 most frequent categories, illustrating a long-tail pattern consistent with open-world scenarios. (c) Breakdown of instruction types, where descriptive instructions dominate to enhance complex reasoning capabilities. (d) Word cloud of descriptive instructions, emphasizing the prevalence of spatial prepositions and attribute adjectives for discriminative target description.}
\label{fig:sa1b-ors-statistic}
\end{figure}

\subsubsection{Distribution of SA1B-ORS}
We conduct a multi-dimensional statistical analysis to characterize the SA1B-ORS dataset. First, as shown in Figure ~\ref{fig:sa1b-ors-statistic}(a), we compare the scale of SA1B-ORS with existing benchmarks. Our dataset significantly surpasses the RefCOCO series and LVIS in both category count and sample size, demonstrating its superior categorical diversity.
Second, we analyze the class distribution. Figure~\ref{fig:sa1b-ors-statistic}(b) visualizes the top 20 most frequent categories, where common classes such as person, building, and sky dominate, followed by a steady decline in instance counts for other classes. This long-tail distribution aligns with real-world visual data patterns in open-world scenarios.
Third, we examine the composition of instruction types. As illustrated in Figure~\ref{fig:sa1b-ors-statistic}(c), SA1B-ORS includes category-based instructions (for single or multiple instance segmentation) and descriptive instructions. Notably, descriptive instructions account for 64.8\% of the data; this prevalence is intentionally designed to enhance the model’s ability to interpret complex referring expressions in challenging environments.
Finally, to understand the linguistic characteristics of these descriptive instructions, we generate a word cloud (Figure~\ref{fig:sa1b-ors-statistic}(d)). The visualization highlights the frequent use of spatial prepositions (e.g., \textit{in}, \textit{above}, \textit{left}, \textit{foreground}) and attribute adjectives (e.g., \textit{blue}, \textit{visible}, \textit{dark},  \textit{far}). These elements work together to provide discriminative descriptions that accurately locate specific instances within complex scenes. Collectively, these statistical properties establish a robust data foundation for open-world referring segmentation.

\subsection{ORS-Bench}

\subsubsection{Overview}

To comprehensively evaluate model performance on the open-world referring segmentation task, we construct a high-quality evaluation suite, \textbf{ORS-Bench}, which comprises \textbf{ORS-ID-Bench} and \textbf{ORS-OOD-Bench}. These two subsets are designed to assess the model's generalization ability under in-distribution (ID) and out-of-distribution (OOD) settings, respectively.

\subsubsection{ORS-ID-Bench: In-Distribution Benchmark}
We first develop an in-distribution benchmark derived from the identical data construction pipeline used during training. It comprises 9,055 high-quality samples distributed across four instruction types: single-instance category instructions (2,465), multiple-instance category instructions (1,823), phrasal instructions (2,946), and descriptive instructions (1,821). The category-based instructions are sourced from the public COCO and LVIS datasets, complemented by our self-constructed SA1B-CoRS dataset. Phrasal instructions are curated from the validation splits of RefCOCO, RefCOCO+, and RefCOCOg, while descriptive instructions originate from our custom-built DeRS dataset. To ensure evaluation reliability, each sample has undergone rigorous manual verification, guaranteeing both the precision of segmentation mask annotations and strict semantic alignment between the referring instructions and their target instances. By encompassing a diverse spectrum of instruction formats with meticulously verified data, this benchmark ensures both the fidelity and representativeness of the evaluation process.

\subsubsection{ORS-OOD-Bench: Out-of-Distribution Benchmark}

\begin{figure}[t]
\centering
\includegraphics[width= 0.9\linewidth]{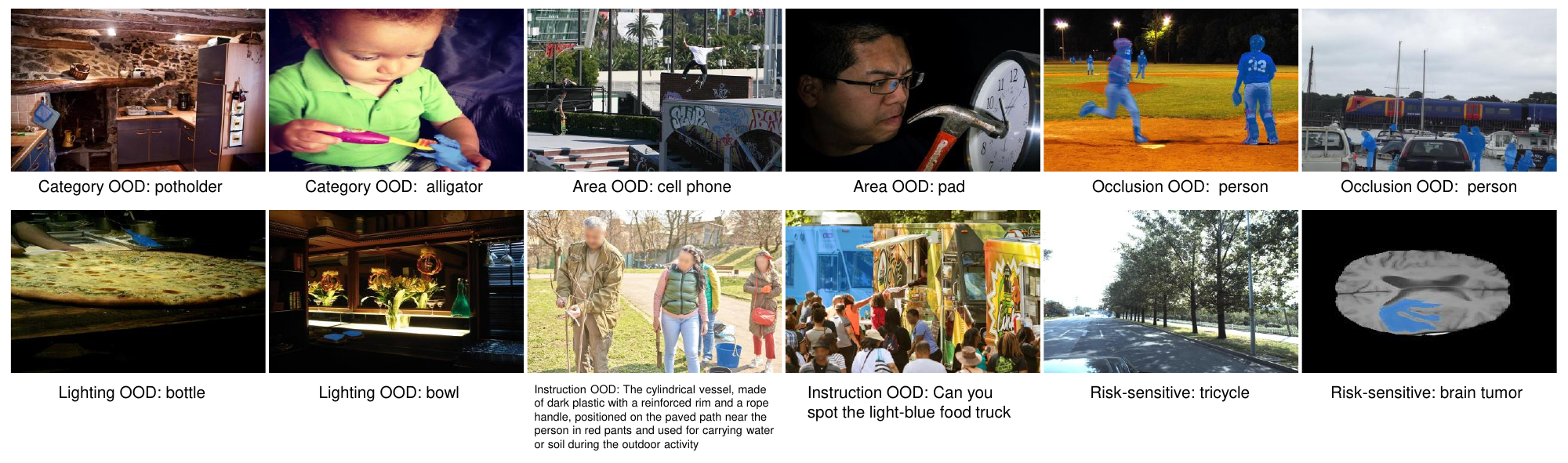}
   \caption{
   Examples from ORS-OOD-Bench. The out-of-distribution benchmark consists of six types of challenge and risk-sensitive scene samples, including {category OOD}, {area OOD}, {occlusion OOD}, {lighting OOD}, {instruction OOD}, and {risk-sensitive scene}.
}
\label{ood_data_vis}
\end{figure}

To further explore the application scope, capability boundaries, and reliability of our model under risk-sensitive scenarios, we construct an out-of-distribution evaluation set, \textbf{ORS-OOD-Bench} (as shown in Figure~\ref{ood_data_vis}). This benchmark comprises data types that are either entirely absent or extremely rare in the training set, enabling the assessment of model generalization in challenging real-world scenarios where objects are difficult to recognize or segment. To curate these challenging samples, we first sample data from the training set and employ an MLLM to annotate attributes across multiple dimensions, thereby characterizing the training data distribution. Specifically, we focus on six key dimensions: target instance category, instance-to-image area ratio, descriptive instruction style, occlusion level, image lighting intensity, and risk-sensitivity. Guided by this distributional analysis, we select samples from a non-overlapping complementary dataset that exhibit significant distributional shifts across these dimensions. These samples include missing or highly underrepresented data types, designated as OOD instances, which undergo rigorous manual verification to ensure the accuracy of both referring instructions and mask annotations.

Through this pipeline, we curate 200 samples for each of the six OOD categories: \textit{category OOD}, \textit{area OOD}, \textit{instruction OOD}, \textit{occlusion OOD}, \textit{lighting OOD}, and \textit{risk-sensitive scene}. Specifically:

\begin{itemize}
    \item \textbf{Category OOD:} We analyze the distribution of object categories in the training set and select categories that are either absent or have a frequency lower than $10^{-4}$ as category OOD examples.
    \item \textbf{Area OOD:} We select instances with extremely small (area ratio $<0.002$) or extremely large (area ratio $>0.7$) spatial extents.
    \item \textbf{Instruction OOD:} We modify the MLLM-based instruction generation process to produce five challenging styles: (1) indirect descriptions that prohibit direct target naming, (2) instructions with negative or exclusionary constraints, (3) interrogative or interactive queries, (4) prompts requiring complex spatial or inter-object relational reasoning, and (5) lengthy descriptions that embed the target within an extensive contextual background.
    \item \textbf{Occlusion OOD:} We prioritize targets with high occlusion ratios.
    \item \textbf{Lighting OOD:} We select images captured under low-light or nighttime conditions.
    \item \textbf{Risk-Sensitive Scene:} We include domains such as autonomous driving and medical diagnosis, which are virtually absent from the training data.
\end{itemize}

We posit that this carefully constructed suite serves as a comprehensive and diverse OOD benchmark for Referring Entity Segmentation. It effectively measures the generalization capabilities and operational boundaries of state-of-the-art referring segmentation models, while providing a standardized testbed to catalyze future research in robust visual grounding.

\section{Implementation}

\subsection{Multi-stage Training}

We adopt a two-stage training strategy to achieve the simultaneous acquisition of robust multimodal understanding and fine-grained perception capabilities for dense prediction tasks. This progression ensures that the model's reasoning logic and spatial awareness are optimized in a synergistic manner.

\paragraph{Stage 1: Segmentation-Centric Adaptation}\label{sec:stage1} 
The primary objective of the first stage is to establish a strong referring segmentation capability by exploiting the grounding proficiency already present in the pretrained vision-language backbone. Since the base model can localize objects from natural-language instructions, we repurpose this spatial prior for dense pixel-level prediction. Concretely, we adapt the language model through LoRA, together with the vision encoder and the mask decoder, under a unified instruction-following framework. Using a fixed input template, the model is trained on a mixture of public referring segmentation datasets and SA1B-ORS. During training, the model is supervised to generate structured outputs that contain bounding boxes and mask placeholders, while the mask decoder maps the corresponding mask tokens to final binary masks. This stage bridges coarse-grained grounding and fine-grained mask prediction, while the simultaneous fine-tuning of the vision encoder further improves mask-oriented visual representations.

\paragraph{Stage 2: Synergistic Enhancement of Perception and Understanding}\label{sec:stage2} 
Upon Stage 1, the second stage aims to recover and strengthen the model's general multimodal understanding and reasoning capabilities while maintaining and further refining referring segmentation performance. We first merge the LoRA weights from Stage 1 into the LLM backbone, and then perform full fine-tuning over the LLM backbone and the mask decoder, while keeping the vision encoder frozen. Training is conducted on a mixed dataset with a ratio of 3:1:2 for referring segmentation, general multimodal understanding, and multimodal reasoning, respectively. For the reasoning portion, we employ the original Qwen3-VL-Instruct model to generate STEM-focused distillation data via off-policy distillation. This stage improves language-centric understanding and reasoning while preserving the visual representations learned in the first stage.

\subsection{Instruction Paradigm and Prompt Design}

To support open-vocabulary capabilities, our instruction-following paradigm strictly adheres to the Qwen3-VL context structure. This approach utilizes the ChatML format, enabling the seamless interleaving of multimodal inputs within a structured, conversational framework.

\paragraph{Instruction Templates}
We define distinct task-specific templates to differentiate between category-level and description-based segmentation. Depending on the granularity of the instruction, the user prompts are formulated as follows:

\begin{itemize}
    \item Category-based Segmentation: 
    \textit{“Locate and segment every instance that belongs to the following categories <category>, report bbox coordinates and masks in JSON format.”}
    
    \item Description-based Segmentation: 
    \textit{“Locate and segment the object that matches the description <description> in the image, report bbox coordinates and masks in JSON format.”}
\end{itemize}

\paragraph{Structured Output Format}
To ensure the output is highly structured and machine-readable, the model is constrained to generate a deterministic JSON schema. For each identified instance, the model predicts three key attributes: \texttt{bbox\_2d} for normalized 2D bounding box coordinates, \texttt{label} for the category name or identifier, and \texttt{mask} for a specific placeholder (e.g., \verb|<mask_start><mask_token><mask_end>|). This placeholder serves as the bridge to the mask decoder for binary mask generation.

\paragraph{Contextual Conversation Example}
The following block illustrates the input and output template used during training and inference for a category-based instruction, following the Qwen3-VL protocol:

\begin{promptblock}{Input Template for Category instruction}
\begin{verbatim}
<|im_start|>system
You are a helpful assistant.<|im_end|>
<|im_start|>user
<|vision_start|><|image_pad|><|vision_end|>
Locate and segment every instance that belongs to the following categories 
"dog", report bbox coordinates and masks in JSON format.<|im_end|>
<|im_start|>assistant
```json
[
  {
    "bbox_2d": [210, 45, 890, 520],
    "label": "dog",
    "mask": "<mask_start><mask_token><mask_end>"
  }
]
```<|im_end|>
<|endoftext|>
\end{verbatim}
\end{promptblock}

\subsection{Implementation Details}

We utilize Qwen3-VL-4B~\citep{Qwen3-VL} as our foundational backbone. The newly introduced mask decoder is lightweight, containing only 17M parameters, which accounts for approximately 0.4\% of the base MLLM model. This minimal overhead enables precise pixel-level mask prediction while largely preserving the efficiency advantage of the original backbone. In the first stage, we employ LoRA~\citep{DBLP:conf/iclr/HuSWALWWC22} with a rank of 32 for 10,000 iterations. During this phase, the original LLM weights are frozen, while the LoRA adapters, the vision encoder, and the mask decoder remain trainable. We employ an initial learning rate of $1\times10^{-4}$ with a cosine annealing schedule. The vision encoder is assigned a $0.01\times$ differential learning rate to facilitate segmentation-specific adaptation while safeguarding the pre-trained feature manifold against over-optimization and distortion. The model is optimized end-to-end using a hybrid objective that combines text generation loss with a segmentation loss, the latter being a weighted sum of per-pixel Binary Cross-Entropy (BCE) loss and DICE loss~\citep{rezatofighi2019generalized}.

In the second stage, the LoRA weights are merged into the LLM backbone, followed by full fine-tuning for 5,000 iterations. In this stage, the LLM backbone and the mask decoder are jointly updated, while the vision encoder remains frozen. We utilize a conservative learning rate of $7\times10^{-7}$ and a cosine decay schedule to perform this final refinement.

\section{Experiments}

\subsection{Experimental Setup}
\paragraph{Datasets.}
We first train on the RefCOCO series~\citep{kazemzadeh2014referitgame,yu2016modeling,mao2016generation} for RES and REC tasks. We then further train the model for open-world referring segmentation using a mixed training set consisting of two public datasets, LVIS~\citep{gupta2019lvis} and COCO~\citep{lin2014microsoft}, together with our constructed SA1B-ORS dataset. For evaluation, we use the validation/test splits of the RefCOCO series for RES and REC, and ORS-Bench for open-world referring segmentation.

\paragraph{Evaluation Metrics.}
For referring segmentation tasks, we use mIoU (the mean of per-sample Intersection-over-Union), cIoU (the overall intersection-over-union computed across the entire dataset)~\citep{xia2024gsva}, and precision@\(t\) (the percentage of samples whose mask IoU exceeds a threshold \(t \in \{0.5, 0.7, 0.9\}\)). For REC, following previous work~\citep{lai2024lisa}, a prediction is considered correct if its IoU with the ground-truth box exceeds 0.5 (Prec@0.5).

The evaluation protocol differs slightly between single-instance and multiple-instance settings. For single-instance segmentation, the metrics are computed directly from the predicted mask and the ground-truth mask. For multiple-instance benchmarks, we first compute a pairwise IoU matrix between all predicted masks and ground-truth masks, and then use the Hungarian algorithm to obtain the optimal one-to-one assignment. The matched prediction--ground-truth pairs are subsequently evaluated with the same metrics as in the single-instance setting.

\input{tables/res.tex}

\subsection{Referring Expression Segmentation}

We evaluate Qwen3-VL-Seg on the standard RefCOCO, RefCOCO+, and RefCOCOg benchmarks using cIoU. As shown in Table~\ref{tab:res}, our model achieves the best results on 6 of the 8 evaluation splits, reaching 82.3/83.7/79.1 on RefCOCO, 76.2/80.2/70.8 on RefCOCO+, and 78.2/78.1 on RefCOCOg.

Compared with SAM-based methods, Qwen3-VL-Seg consistently delivers higher mask quality without relying on an external segmentation model. For example, it surpasses LISA by 7.4 points on RefCOCO Val and 12.7 points on RefCOCO+ TestB, and exceeds SegAgent+SAM by 3.1 points and 5.4 points on the same splits, respectively. The comparison with SAM3 is also notable: Qwen3-VL-Seg improves from 63.4 to 70.8 on RefCOCO+ TestB and from 74.0 to 78.1 on RefCOCOg Test, suggesting that for referring expression segmentation, MLLM-based grounding and language understanding remain more decisive than the generic concept-segmentation capability of pure segmentation foundation models.

Among SAM-free alternatives, Qwen3-VL-Seg remains highly competitive across all benchmarks and outperforms Unipixel, which also adopts a Qwen-family backbone, by 1.8 points on RefCOCO Val and 1.3 points on RefCOCO+ TestA. These gains indicate that explicitly treating the predicted box as a structural prior provides more effective spatial constraints than a fully unified pixel-prediction formulation. Moreover, Qwen3-VL-Seg remains more accurate and substantially more parameter-efficient than 8B models such as UFO and Text4Seg, demonstrating that the proposed box-guided mask decoder can recover high-quality masks within a compact 4B framework.

\input{tables/rec.tex}

\subsection{Referring Expression Comprehension.}
As summarized in Table \ref{tab:rec}, Qwen3-VL-Seg achieves state-of-the-art performance among all MLLM-based RES methods while remaining highly competitive against specialized Vision Generalist Models and large-scale MLLMs. Compared to \textit{Vision Generalists}, our model consistently outperforms Grounding DINO across all benchmarks and narrows the performance gap with Florence-2 to a marginal difference. Within the \textit{General-purpose MLLMs} category, Qwen3-VL-Seg exhibits robust grounding proficiency, surpassing InternVL-3.5 on several key splits and following closely behind the state-of-the-art Youtu-VL.

Notably, Qwen3-VL-Seg shows a significant performance leap over its foundational backbone, Qwen3-VL, across all benchmarks. While the backbone provides the essential grounding capability, our results indicate that the referring segmentation task acts as a powerful dense supervisor that further sharpens spatial accuracy. Specifically, we observe a consistent improvement in Prec@0.5, such as the increase from 90.7\% to 92.7\% on RefCOCO Val and a substantial 6.6\% jump on RefCOCO+ TestB. This suggests that pixel-level supervision encourages a tighter semantic-spatial alignment, effectively refining the backbone’s inherent grounding ability through more granular feature optimization.

\begin{figure}[t]
\centering
\includegraphics[width=0.9\linewidth]{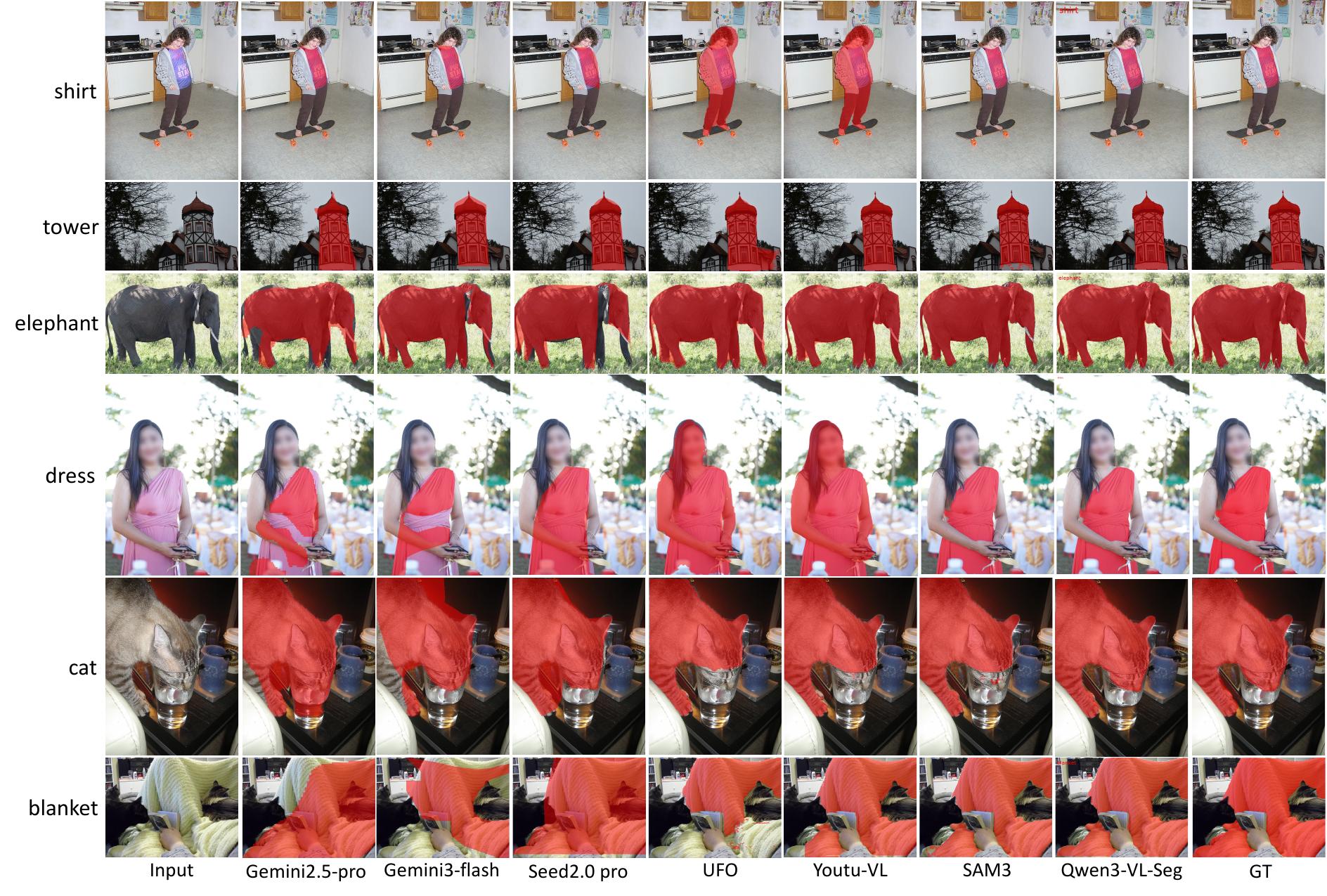}
\caption{Qualitative comparison results of open-world referring segmentation with single-instance category instructions. The predicted mask is overlayed on the input image in red for visualization.}
\label{fig:vis_singleins}
\end{figure}

\subsection{Open-world Referring Segmentation}

\paragraph{Quantitative Evaluation.}

\input{tables/bench.tex}

\begin{figure}[h]
\centering
\includegraphics[width=1\linewidth]{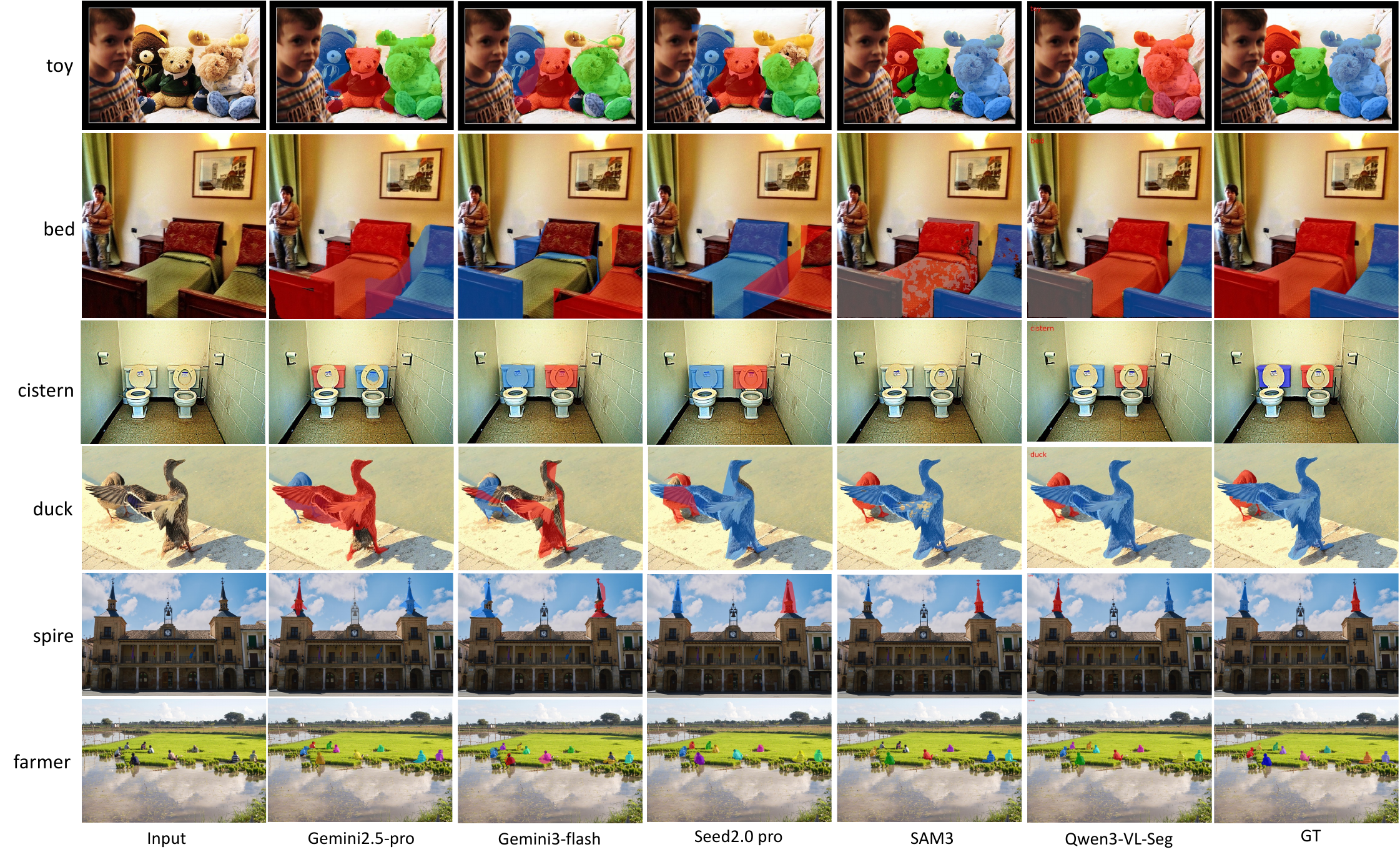}
\caption{Qualitative comparison results of open-world referring segmentation with multiple-instance category instructions. The predicted multiple masks are overlayed on the input image with different colors for visualization.}
\label{fig:vis_multiins}
\end{figure}

Table~\ref{tab:bench_cmp} reports results on ORS-ID-Bench, which evaluates in-distribution open-world referring segmentation under four instruction formats: single-instance category instructions, multiple-instance category instructions, phrasal instructions, and descriptive instructions. Qwen3-VL-Seg achieves the best performance in 7 out of 8 metrics, reaching 96.0/97.4 on single-instance category instructions, 93.2/91.6 on multiple-instance category instructions, 82.8/94.3 on phrasal instructions, and 91.0/91.3 on descriptive instructions in terms of cIoU/P@0.5. These results indicate that the proposed framework remains strong not only on category-oriented grounding but also on more flexible referring formats that require tighter alignment between language understanding and mask prediction.

In category-oriented settings, SAM3 is the strongest competing baseline, achieving 94.2 cIoU and 97.3 P@0.5 on single-instance instructions, and 94.6 cIoU and 91.4 P@0.5 on multiple-instance instructions. Qwen3-VL-Seg outperforms SAM3 on all single-instance metrics and attains the best P@0.5 in the multiple-instance setting, while trailing slightly in multiple-instance cIoU by 1.4 points. This comparison suggests that large segmentation foundation models remain highly competitive when the instruction is closely aligned with visual concepts, but our box-guided MLLM-native design can match or exceed their precision without introducing an external segmenter.

The more decisive advantage of Qwen3-VL-Seg appears in language-intensive settings. On phrasal instructions, our model surpasses the best baseline by 19.0 cIoU points and 13.6 P@0.5 points. On descriptive instructions, it further improves over the strongest baseline by 15.5 cIoU points and 13.1 P@0.5 points. In contrast, both general-purpose MLLMs and segmentation-centric models exhibit a pronounced performance drop once the target can no longer be specified by a simple category name alone. We attribute this gain to the fact that Qwen3-VL-Seg directly converts MLLM grounding priors into dense masks, allowing linguistic semantics, box-level localization, and pixel-level refinement to be optimized within a unified framework.

\paragraph{Qualitative Evaluation.} Figure~\ref{fig:vis_singleins} and Figure~\ref{fig:vis_multiins} present qualitative comparisons on single-instance and multiple-instance category instructions, respectively. We compare three distinct architectural paradigms: general-purpose Multimodal Large Language Models (MLLMs) such as Gemini-2.5-pro, Gemini-3-flash, and Seed-2.0 Pro; unified generalist perception models such as UFO and Youtu-VL; and the specialized segmentation foundation model SAM3. Since UFO and Youtu-VL do not support multiple-instance segmentation, we omit them from Figure~\ref{fig:vis_multiins}. While the general-purpose MLLMs demonstrate reliable semantic localization, their predicted masks often lack the pixel-level resolution required to delineate complex boundaries, as illustrated by cases such as the elephant and blanket examples for Gemini and Seed. In contrast, although unified generalist models and specialized segmentation frameworks often produce higher-quality masks, they may still exhibit inconsistencies in instance coverage or semantic grounding under diverse open-world category instructions, as shown by cases such as shirt for UFO and Youtu-VL, and cistern and farmer for SAM3. Qwen3-VL-Seg combines the grounding capability of a strong vision-language backbone with the boundary recovery capacity of the proposed box-guided mask decoder. As a result, its predicted masks remain semantically aligned with the referring instruction while preserving accurate object geometry and boundary detail across diverse scenarios. This combination yields consistently stronger robustness and fidelity than both unified generalist architectures and specialized segmentation models.

\begin{figure}[t]
\centering
\includegraphics[width=\linewidth]{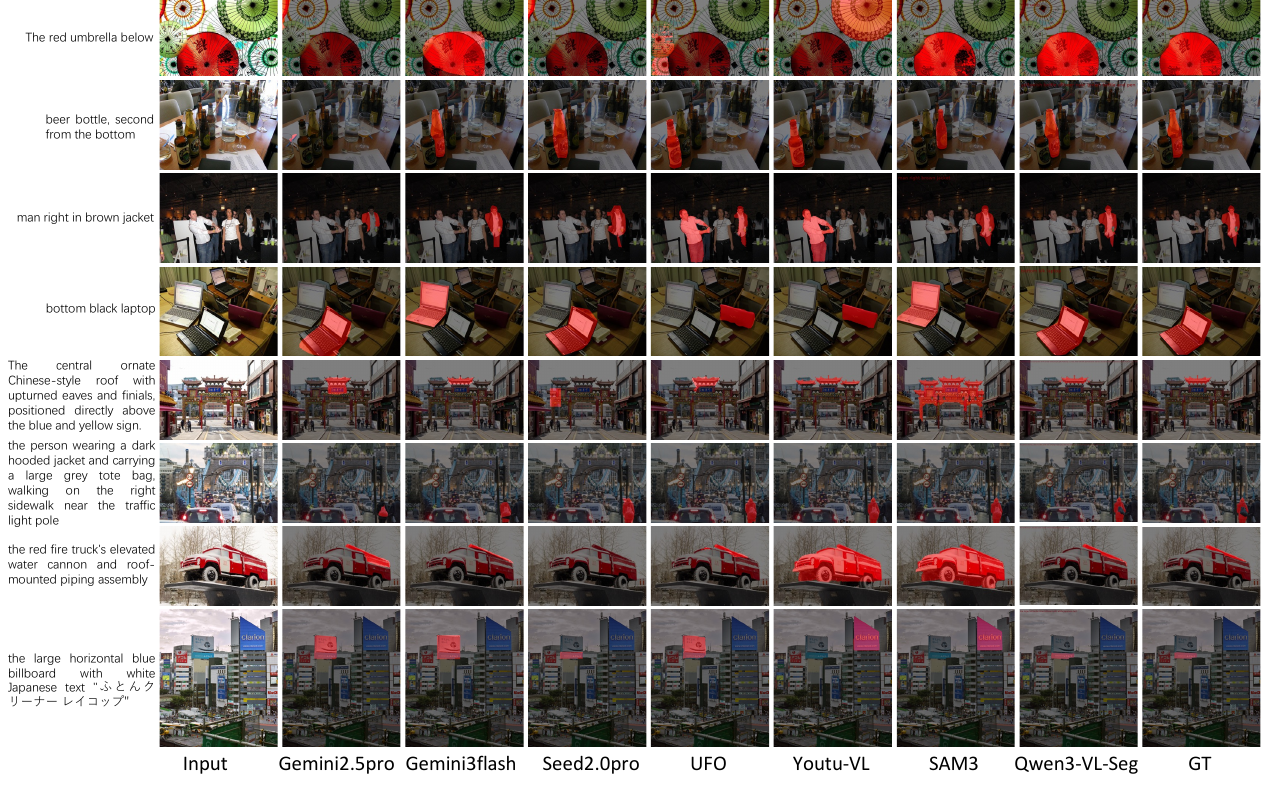}
\caption{Qualitative comparison results of open-world referring segmentation with phrasal and descriptive instructions. The predicted mask is overlayed on the input image in red for visualization.}
\label{fig:vis_des}
\end{figure}

Figure~\ref{fig:vis_des} further presents qualitative results on phrasal and descriptive instructions. The visualizations show that Qwen3-VL-Seg excels at both precise spatial reasoning and fine-grained part-level grounding, effectively resolving complex referring instructions that remain challenging for existing models. Competing methods often struggle with relative positioning or sub-object-level descriptions, such as identifying the "second bottle from the bottom" or isolating the "elevated water cannon" of a fire truck, whereas our model consistently achieves accurate localization and sharp pixel-level segmentation. By combining robust linguistic understanding with the proposed box-guided mask decoder, Qwen3-VL-Seg produces masks that are both semantically grounded and geometrically precise, even for highly localized targets in cluttered open-world scenes.

\subsection{OOD Referring Segmentation}

\begin{figure}[t]
\centering
\includegraphics[width= 0.8\linewidth]{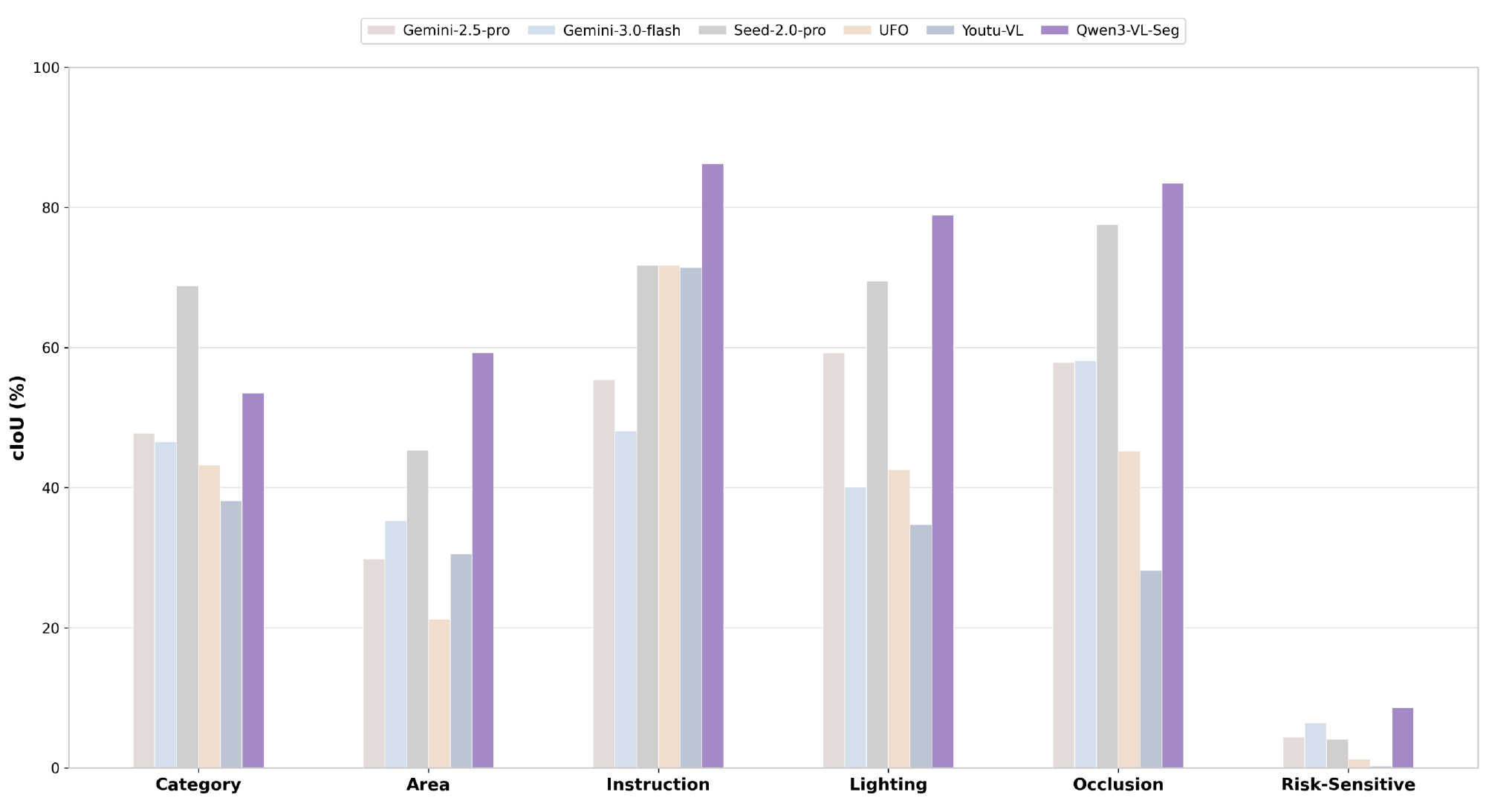}
   \caption{
   Comparison of OOD Referring Segmentation on ORS-OOD-Bench.
}
\label{fig:ood_comparison}
\end{figure}

To evaluate the generalization capability of Qwen3-VL-Seg in challenging scenarios, we conduct a comprehensive assessment and comparison of various methods on our constructed ORS-OOD-Bench benchmark. As illustrated in Figure~\ref{fig:ood_comparison}, Qwen3-VL-Seg achieves cIoU scores of 53.49\%, 59.30\%, 86.22\%, 78.9\%, 83.45\%, and 8.64\% across six categories of OOD scenarios, including category, area, instruction, lighting, occlusion, and risk-sensitive scenes, respectively. These results significantly outperform other frontier MLLMs and unified generalist perception models. This demonstrates that our approach retains superior generalization capabilities, even when encountering novel, unseen scenarios and challenging, difficult data samples.

Despite outperforming current state-of-the-art methods in OOD settings, all methods exhibit a notable performance drop compared to their results on the ORS-ID-Bench. This decline is particularly pronounced in risk-sensitive scenes, such as autonomous driving and medical diagnosis, where nearly all methods failed to achieve accurate identification and segmentation. This observation highlights that future referring segmentation models must further enhance their out-of-distribution generalization capabilities to continuously expand the boundaries of model adaptability for diverse application scenarios.

\input{tables/vl_comparisons.tex}

\subsection{General Multimodal Evaluation}

To further assess how the proposed multi-stage training strategy affects general-purpose multimodal competence, we evaluate the original Qwen3-VL-4B instruct model together with its stage-1 and stage-2 variants on a diverse suite of benchmarks spanning visual question answering, multimodal reasoning, OCR-intensive understanding, and visual grounding. Table~\ref{tab:multimodal_comparison} reveals a stage-wise trend that closely mirrors the training objectives described in Section~\ref{sec:stage1} and Section~\ref{sec:stage2}. After the segmentation-centric adaptation of Stage 1, the model improves on perception-oriented tasks such as MMBench-EN and RefCOCO$_{\text{val}}$, but exhibits broader regressions on several reasoning- and OCR-intensive benchmarks, indicating that dense segmentation supervision and vision-encoder adaptation shift the model toward stronger spatial specialization. The second stage largely restores this balance. By reintroducing general multimodal understanding and reasoning data while keeping the vision encoder frozen, the final Qwen3-VL-Seg model recovers markedly over the stage-1 variant on MMStar, MMMU$_{\text{val}}$, MathVision, MathVista$_{\text{mini}}$, InfoVQA$_{\text{val}}$, DocVQA$_{\text{val}}$, CharXiv$_{\text{RQ}}$, RealWorldQA, and RefCOCO$_{\text{val}}$. It remains particularly strong on OCR- and grounding-related tasks, achieving 81.4 on InfoVQA$_{\text{val}}$, 45.2 on CharXiv$_{\text{RQ}}$, 71.2 on RealWorldQA, and 92.3 on RefCOCO$_{\text{val}}$, which are superior to the compared 4B-scale baselines. Overall, these results suggest that the proposed two-stage recipe yields a favorable rebalancing of model capabilities, with the first stage injecting segmentation-oriented spatial specialization and the second stage restoring a stronger balance between fine-grained perception and general multimodal understanding.

\subsection{Ablation studies}

\input{tables/ablation.tex}

\paragraph{Ablation on segmentation architecture.}
Table~\ref{tab:ablation_study} presents an ablation study of our segmentation architecture on the RefCOCO, RefCOCO+, and RefCOCOg validation sets. Compared with the \emph{Qwen box+SAM} baseline, both the full model and its ablated variants achieve substantial gains across all metrics, confirming the effectiveness of the proposed box-guided decoder in converting MLLM grounding outputs into accurate pixel-level masks. Notably, the full model improves the strict P@0.9 metric from 39.6 to 50.2 on RefCOCO, from 36.7 to 47.6 on RefCOCO+, and from 35.9 to 47.3 on RefCOCOg.

The ablated variants further clarify the contribution of each component. Freezing the vision encoder (\emph{Ours (freeze vit)}) consistently degrades performance, particularly on the high-threshold P@0.9 metric (e.g., a 6.1-point drop on RefCOCO), indicating that adapting the visual backbone is important for learning segmentation-oriented representations beyond coarse grounding. Similarly, removing multi-scale visual features (\emph{Ours (w/o multivit)}) leads to consistent drops in cIoU and P@0.9, suggesting that intermediate visual features provide useful spatial detail beyond the top-layer representation alone.

The shallow image branch also consistently contributes: removing it (\emph{Ours (w/o image)}) hurts performance across all three datasets, showing that direct high-resolution image cues help recover more precise object boundaries. Notably, the performance gap between the full model and the ablated variants is most evident on P@0.9, which is the most sensitive metric to boundary quality. This suggests that, while coarse localization can already be achieved with fewer components, multi-scale feature injection and the high-resolution image branch are particularly important for improving mask precision and completeness. Overall, the results show that the proposed components are complementary and that their combination yields the most consistent segmentation performance.

\section{Conclusion}

We present Qwen3-VL-Seg, a lightweight framework for open-world referring segmentation that extends the pretrained MLLM from box-level grounding to pixel-level mask prediction. The key idea is to treat the MLLM-predicted box as a structural prior for mask decoding, enabling precise segmentation without relying on external segmentation foundation models. To support scalable training and evaluation, we construct the SA1B-ORS training dataset, with complementary category-oriented and descriptive subsets, and introduce ORS-Bench for in-distribution and out-of-distribution assessment of open-world referring segmentation. Experiments on referring expression segmentation, visual grounding, and open-world benchmarks indicate that Qwen3-VL-Seg performs strongly across both closed-set and open-world settings, with especially clear gains on language-intensive instructions and strong out-of-distribution generalization. Additional general multimodal evaluation shows that the model broadly preserves general-purpose multimodal competence after segmentation-oriented adaptation. These findings suggest that reusing MLLM grounding outputs as structural priors provides an effective path toward unified and efficient referring segmentation.

\bibliography{colm2024_conference}
\bibliographystyle{colm2024_conference}




\end{document}

%% file: tables/res.tex
\begin{table*}[h]
\centering
    \caption{\textbf{Referring Expression Segmentation} results (cIoU) on RefCOCO (+/g) datasets. The models we compared are all MLLM-based methods and the best results are marked in bold.} 
    \setlength{\tabcolsep}{5pt}
    \begin{tabular}{l|ccc|ccc|cc}
    \hline
    \multirow{2}*{Methods} & \multicolumn{3}{c|}{RefCOCO} & \multicolumn{3}{c|}{RefCOCO+} & \multicolumn{2}{c}{RefCOCOg} \\
    \cline{2-9}
     &  Val & TestA & TestB & Val & TestA & TestB & Val & Test \\
    \hline
    LISA$_{\text{LLaVA-7B}}$(ft)\citep{lai2024lisa}          & 74.9 & 79.1 & 72.3 & 65.1 & 70.8 & 58.1 & 67.9 & 70.6  \\
    GSVA$_{\text{Vicuna-7B}}$(ft)\citep{xia2024gsva}          & 77.2 & 78.9 & 73.5 & 65.9 & 69.6 & 59.8 & 72.7 & 73.3  \\
    AnyRef$_{\text{LLaVA-7B}}$(ft)\citep{he2024multi}     & 76.9 & 79.9 & 74.2 & 70.3 & 73.5 & 61.8 & 70.0 & 70.7  \\
    PixelLM$_{\text{LLaVA-7B}}$\citep{ren2024pixellm}     & 73.0 &  76.5 &  68.2 &  66.3 &  71.7 &  58.3 &  69.3 &  70.5  \\
    VisionLLM v2$_{\text{Vicuna-7B}}$\citep{wu2024visionllm}  & 76.6 & 79.3 & 74.3 & 64.5 & 69.8 & 61.5 & 70.7 & 71.2  \\
    Text4Seg$_{\text{InternVL2-8B}}$\citep{lan2024text4seg}  & 79.2 & 81.7 & 75.6 & 72.8 & 77.9 & 66.5 & 74.0 & 75.3  \\
    SegAgent+SAM$_{\text{LLaVA-7B}}$\citep{zhu2025segagent}    & 79.2 & 81.4 & 75.7 & 71.5 & 76.7 & 65.4 & 74.8 & 74.9  \\
    M²SA$_{\text{LLaVA-7B}}$\citep{jang2025mmr}  & 74.0 & 76.8 & 69.7 & 63.1 & 67.2 & 56.1 & 67.0 & 68.3  \\
    UFO$_{\text{InternVL2-8B}}$\citep{tang2025ufo}  & 78.0 & 79.7 & 75.6 & 72.3 & 76.8 & 66.6 & 73.7 & 74.3   \\
    MLLMSeg$_{\text{InternVL2.5-4B}}$\citep{wang2025unlocking}     & 79.5 & 81.4  & 77.7 & 75.4 & 78.7 & 70.8 & 78.1 & \textbf{78.4}  \\
    SAM3~\citep{carion2025sam} & 75.5 & 77.6  & 71.0 & 67.3 & 71.1 & 63.4 & 73.4 & 74.0  \\
    Unipixel$_{\text{Qwen2.5-VL-3B}}$  & 80.5 & 82.6  & 76.9 & 74.3 & 78.9 & 68.4 & 76.3 & 77.0 \\
    Youtu-VL$_{\text{4B}}$\citep{wang2025unlocking}     & 80.7 & 82.0  & 78.4 & 76.2 & 79.6  & \textbf{71.4} & 76.5 & 76.6 \\
    Qwen3-VL-Seg$_{4B}$ (Ours)   & \textbf{82.3}  & \textbf{83.7}  & \textbf{79.1} & \textbf{76.2} & \textbf{80.2} & 70.8 & \textbf{78.2} & 78.1  \\
    \hline
    \end{tabular}
    \label{tab:res}
\end{table*}

%% file: tables/rec.tex
\begin{table*}[h]
\centering
    \caption{\textbf{Referring Expression Comprehension} results (Prec@0.5) on RefCOCO (+/g) datasets. The table compares Vision Generalist Models, General-purpose MLLMs, and existing MLLM-based RES methods. Best results in each setting are marked in bold.} 
    \setlength{\tabcolsep}{5pt}
    \begin{tabular}{l|ccc|ccc|cc}
    \hline
    \multirow{2}*{Methods} & \multicolumn{3}{c|}{RefCOCO} & \multicolumn{3}{c|}{RefCOCO+} & \multicolumn{2}{c}{RefCOCOg} \\
    \cline{2-9}
     &  Val & TestA & TestB & Val & TestA & TestB & Val & Test  \\
    \hline
    \multicolumn{9}{l}{\textit{\textbf{Vision Generalist Models}}} \\
    Florence-2~\citep{xiao2024florence}  & \textbf{93.4} & \textbf{95.3} & \textbf{92.0} & \textbf{88.3} & \textbf{92.9} & \textbf{83.6} & \textbf{91.2} & \textbf{91.7} \\
    Grounding DINO~\citep{liu2023grounding}  & 90.6 & 93.2 & 88.2 & 82.8 & 89.0 & 75.9 & 86.1 & 87.0  \\
    \hline    
    \multicolumn{9}{l}{\textit{\textbf{General-purpose MLLMs}}} \\
    InternVL-3.5$_{4B}$\citep{wang2025internvl3}   & 92.5 & 94.3 & 88.2 & 87.6 & 92.3 & 81.6 & 89.6 & 89.3  \\
    Qwen2.5-VL$_{3B}$\citep{Qwen3-VL}  & 89.1 & 91.7 & 84.0 & 82.4 & 88.0 & 74.1 & 85.2 & 85.7  \\
    Qwen3-VL$_{4B}$\citep{Qwen3-VL}  & 90.7 & 92.2 & 86.7 & 82.9 & 89.4 & 75.6 & 87.3 & 87.7 \\
    Youtu-VL$_{4B}$\citep{wei2026youtu} & \textbf{93.6}&\textbf{95.2}&\textbf{90.8}&\textbf{90.1}&\textbf{93.9}&\textbf{85.4}&\textbf{92.2}&\textbf{92.9} \\
    \hline
    \multicolumn{9}{l}{\textit{\textbf{MLLM-based RES methods}}} \\
    LISA$_{\text{LLaVA-7B}}$(ft)\citep{lai2024lisa}         & 85.4 & 88.8 & 82.6 & 74.2 & 79.5 & 68.4 & 79.3 & 80.4  \\
    GSVA$_{\text{Vicuna-7B}}$(ft)\citep{xia2024gsva}          & 86.3 & 89.2 & 83.8 & 72.8 & 78.8 & 68.0 & 81.6 & 81.8 \\
    PixelLM$_{\text{LLaVA-7B}}$\citep{ren2024pixellm}     & 89.8 & 92.2 & 86.4 & 83.2 & 87.0 & 78.9 & 84.6 & 86.0 \\
    VisionLLM v2$_{\text{Vicuna-7B}}$\citep{wu2024visionllm}  & 87.9 & 91.2 & 84.3 & 77.6 & 83.8 & 70.2 & 82.9 & 84.1\\
    Text4Seg$_{\text{InternVL2-8B}}$\citep{lan2024text4seg}  & 90.3 & 93.4 & 87.5 & 85.2 & 89.9 & 79.5 & 85.4 & 85.4 \\
    UFO$_{\text{InternVL2-8B}}$\citep{tang2025ufo}  & 91.4  & 93.8  & 88.2  & 85.7  & 90.7  & 79.7  & 86.8  & 87.4 \\
    MLLMSeg$_{\text{InternVL2.5-4B}}$\citep{wang2025unlocking}     & 91.6 & 93.3  & 87.6 & 86.0 & 90.0 & 79.8 & 88.3 & 88.6  \\
    Qwen3-VL-Seg$_{4B}$ (Ours)   & \textbf{92.7}  & \textbf{94.6}  & \textbf{89.8} & \textbf{87.8} & \textbf{92.1} & \textbf{82.2} & \textbf{89.1} & \textbf{89.0}  \\
    \hline
    \end{tabular}
    \label{tab:rec}
\end{table*}

%% file: tables/bench.tex
\begin{table}[t]
  \centering
  \caption{\textbf{Open-world Referring Segmentation} results (cIoU and precision@0.5) on the proposed ORS-ID-Bench. Best results in each setting are marked in bold.}
  \label{tab:bench_cmp}
  \setlength{\tabcolsep}{4pt} 
  \small
  \begin{tabular}{lcccccccc}
    \toprule
    \multirow{2}{*}{Methods} & 
    \multicolumn{2}{c}{Single-instance} & 
    \multicolumn{2}{c}{Multiple-instance} & 
    \multicolumn{2}{c}{Phrasal} & 
    \multicolumn{2}{c}{Descriptive} \\
    
     & 
    \multicolumn{2}{c}{Category Instruct} & 
    \multicolumn{2}{c}{Category Instruct} & 
    \multicolumn{2}{c}{Instruct} & 
    \multicolumn{2}{c}{Instruct} \\
    
    \cmidrule(lr){2-3} \cmidrule(lr){4-5} \cmidrule(lr){6-7} \cmidrule(lr){8-9}
     & cIoU & p@0.5 & cIoU & p@0.5 & cIoU & p@0.5 & cIoU & p@0.5 \\
    \midrule
    
    Gemini-2.5-pro~\citep{comanici2025gemini} & 77.7 & 85.1 & 75.9 & 83.8 & 58.7 & 73.1 & 44.6 & 45.4 \\
    Gemini-3.0-flash~\citep{gemini3_flash_model_card} & 67.3 & 71.1 & 62.1 & 58.5 & 51.6 & 61.2 & 57.2 & 70.9 \\
    Seed-2.0-pro~\citep{seed2_tech_report}   & 65.4 & 69.9 & 66.4 & 73.0 & 63.8 & 80.7 & 61.0 & 74.7 \\
    Youtu-VL$_{\text{4B}}$\citep{wang2025unlocking}      & 77.6 & 86.4 & -    & -    & 53.2 & 69.0 & 69.3 & 76.0 \\
    UFO$_{\text{InternVL2.5-8B}}$\citep{tang2025ufo}            & 72.4 & 83.7 & -    & -    & 52.1 & 67.2 & 73.7 & 70.9 \\
    SAM3~\citep{carion2025sam}           & 94.2 & 97.3 & \textbf{94.6} & 91.4 & 66.6 & 79.1 & 75.5 & 78.2 \\
    \textbf{Qwen3-VL-Seg} & \textbf{96.0} & \textbf{97.4} & 93.2 & \textbf{91.6} & \textbf{82.8} & \textbf{94.3} & \textbf{91.0} & \textbf{91.3} \\
    
    \bottomrule
  \end{tabular}
\end{table}

%% file: tables/vl_comparisons.tex
\begin{table}[t]
\centering
\caption{Result comparison of Qwen3-VL-Seg with the existing MLLMs on various general-purpose multimodal benchmarks}
\label{tab:multimodal_comparison}
\resizebox{0.9\textwidth}{!}{%
\begin{tabular}{l c c c c}
\hline
\textbf{Benchmarks} & \makecell{InternVL-3.5\\4B} & \makecell{Qwen3-VL\\4B (instruct)} & \makecell{Qwen3-VL\\4B (S-1)} & \makecell{Qwen3-VL-Seg\\4B (S-2)} \\
\hline
\textit{\textbf{General VQA}} &  &  &  & \\
MMStar & 65.0 & \textbf{69.8}  & 67.5 & 67.7 \\
MMBench-EN & 80.3 & 83.9 & \textbf{86.2} & 84.2 \\
\hline
\textit{\textbf{Multimodal Reasoning $\&$ Math}} & & & & \\
MMMU$_{\text{val}}$ & 66.6 & \textbf{67.4} & 63.4 & 66.2 \\
MMMU-Pro & --  & \textbf{53.2} & 51.5 & 51.3 \\
MathVision & -- & \textbf{51.6} & 47.9 & 50.4 \\
MathVista$_{\text{mini}}$ & \textbf{77.1} & 73.7 & 70.9 & 75.5 \\
\hline
\textit{\textbf{OCR-related Understanding}} & & & & \\
AI2D$_{\text{test}}$ & 82.6 & \textbf{84.1} & 81.6 & 79.7 \\
InfoVQA$_{\text{val}}$ & 78.0 & 80.3 & 75.2 & \textbf{81.4} \\
DocVQA$_{\text{val}}$ & 92.4  & \textbf{95.3} & 93.8 & 94.1 \\
CharXiv$_{\text{RQ}}$ & 39.6 & 39.7 & 38.9 & \textbf{45.2} \\
RealWorldQA & 66.3 & 70.9 & 68.0 & \textbf{71.2} \\
\hline
\textit{\textbf{Visual Grounding}} & & & & \\
RefCOCO$_{\text{val}}$ & \textbf{92.5} & 91.6 & 91.8 & 92.3 \\
\hline
\end{tabular}%
}
\end{table}

%% file: tables/ablation.tex
\begin{table}[ht]
\centering
\caption{Ablation experiments on RefCOCO, RefCOCO+, and RefCOCOg (VAL) datasets.}
\label{tab:ablation_study}
\begin{tabular}{llccccc}
\toprule
\textbf{Dataset} & \textbf{Method} & \textbf{mIoU} & \textbf{cIoU} & \textbf{P@0.5 (\%)} & \textbf{P@0.7 (\%)} & \textbf{P@0.9 (\%)} \\ \midrule
\multirow{5}{*}{\begin{tabular}[c]{@{}l@{}}RefCOCO\\ (VAL)\end{tabular}} 
 & Qwen box+SAM & 74.3 & 70.3 & 83.2 & 76.9 & 39.6 \\
 & Ours (freeze vit) & 81.8 & 80.9 & 92.8 & 87.0 & 44.1 \\
 & Ours (w/o multivit) & 82.7 & 81.9 & 92.7 & 88.7 & 49.0 \\
 & Ours (w/o image) & 82.7 & 81.9 & 92.8 & 88.7 & 48.0 \\
 & \textbf{Ours} & \textbf{82.8} & \textbf{82.3} & \textbf{92.8} & \textbf{88.7} & \textbf{50.2} \\ \midrule
\multirow{5}{*}{\begin{tabular}[c]{@{}l@{}}RefCOCO+\\ (VAL)\end{tabular}} 
 & Qwen box+SAM & 69.3 & 66.0 & 76.9 & 70.8 & 36.7 \\
 & Ours (freeze vit) & 78.5 & 75.5 & 88.3 & 83.0 & 42.5 \\
 & Ours (w/o multivit) & 77.7 & 75.4 & 87.2 & 83.6 & 46.2 \\
 & Ours (w/o image) & 77.5 & 75.3 & 87.1 & 83.5 & 46.1 \\
 & \textbf{Ours} & \textbf{78.5} & \textbf{76.0} & \textbf{87.8} & \textbf{84.1} & \textbf{47.6} \\ \midrule
\multirow{5}{*}{\begin{tabular}[c]{@{}l@{}}RefCOCOg\\ (VAL)\end{tabular}} 
 & Qwen box+SAM & 72.9 & 71.0 & 81.6 & 72.8 & 35.9 \\
 & Ours (freeze vit) & 78.3 & \textbf{78.4} & 87.7 & 81.2 & 43.8 \\
 & Ours (w/o multivit) & 78.1 & 77.4 & 86.8 & 81.5 & 47.1 \\
 & Ours (w/o image) & 77.8 & 76.8 & 87.0 & 81.0 & 46.1 \\
 & \textbf{Ours} & \textbf{78.6} & 78.2 & \textbf{88.0} & \textbf{82.6} & \textbf{47.3} \\ \bottomrule
\end{tabular}
\end{table}